%% file: main.tex
\documentclass[10pt,twocolumn,letterpaper]{article}

\usepackage{cvpr}              
\usepackage{listings}
\usepackage{subcaption}

\input{preamble}

\definecolor{cvprblue}{rgb}{0.21,0.49,0.74}
\usepackage[pagebackref,breaklinks,colorlinks,allcolors=cvprblue]{hyperref}

\def\confName{CVPR}
\def\confYear{2026}

\title{$\beta$-CLIP: Text-Conditioned Contrastive Learning for Multi-Granular \\
Vision-Language Alignment}

\author{
Fatimah Zohra \quad
Chen Zhao \quad
Hani Itani \quad
Bernard Ghanem \\
{King Abdullah University of Science and Technology (KAUST)} 
}

\begin{document}
\maketitle
\input{sec/0_abstract}    
\input{sec/1_intro}
\input{sec/2_relatedwork}

\input{sec/3_method}
\input{sec/4_experiments}
\input{sec/5_ablations}
\input{sec/6_conclusion}

{
    \small
    \bibliographystyle{ieeenat_fullname}
    \bibliography{main}
}
\input{sec/X_suppl}

\end{document}

%% file: preamble.tex
\usepackage{tabularray}
\usepackage{booktabs}
\usepackage{multirow}



%% file: sec/0_abstract.tex
\begin{abstract}
CLIP achieves strong zero-shot image-text retrieval by aligning global vision and text representations, yet it falls behind on fine-grained tasks even when fine-tuned on long, detailed captions. In this work, we propose $\beta$-CLIP, a multi-granular text-conditioned contrastive learning framework designed to achieve hierarchical alignment between multiple textual granularities—from full captions to sentences and phrases—and their corresponding visual regions. For each level of granularity, $\beta$-CLIP utilizes cross-attention to dynamically pool image patches, producing contextualized visual embeddings. To address the semantic overlap inherent in this hierarchy, we introduce the $\beta$-Contextualized Contrastive Alignment Loss ($\beta$-CAL). This objective parameterizes the trade-off between strict query-specific matching and relaxed intra-image contextualization, supporting both soft Cross-Entropy and hard Binary Cross-Entropy formulations. We find that each loss interacts differently with hierarchical supervision: CE's softmax sharpens fine-grained discrimination, while BCE's sigmoid favors long-text retrieval while both benefit from hierarchy. Through extensive experiments, we demonstrate that $\beta$-CLIP significantly improves dense alignment: achieving \textbf{91.8\%} T2I \textbf{92.3\%} I2T at R@1 on Urban1K and \textbf{30.9\%} on FG-OVD (Hard), setting state-of-the-art among methods trained without hard negatives. $\beta$-CLIP establishes a robust, adaptive baseline for dense vision–language correspondence. The code and models are released at \href{https://github.com/fzohra/B-CLIP}{https://github.com/fzohra/B-CLIP}.
\end{abstract}

%% file: sec/1_intro.tex
\input{figs/visualizations}
\section{Introduction}
Vision-language models such as CLIP \cite{clip} have transformed multimodal learning by aligning visual and textual representations in a shared latent space, enabling zero-shot capabilities across diverse tasks such as retrieval, classification, detection, and segmentation in various domains \citep{liang2023ovseg, vild, glip, zhai2022lit, ila, max-pooling}. More recently, CLIP has been widely adopted as either a vision or text backbone in generative models for tasks including captioning, visual question answering (VQA), and even image generation \cite{liu2023llava, molmo2024, zhang2025videollama3, sun2024emugenerative,podell2023sdxl,tang2025unilip}. However, in the era of long texts, CLIP faces two primary bottlenecks: a fixed context length of only 77 tokens and a contrastive training objective that captures coarse, global alignments between entire images and captions~\cite{zhang2024longclip,wu2024lotlip}. Although this global alignment effectively captures broad semantics, and to some extent fine-grained semantics as well \cite{bianchi2024clip}, \textit{it provides no direct mechanism to associate specific visual regions with fine-grained text}. Extending CLIP to accommodate expansive contexts and granular alignments remains a significant challenge.

Prior efforts to address CLIP's coarse-grained limitations have focused on localizing visual concepts in increasingly abstract ways. The most direct approaches explicitly align region-level features with text, for example through early works like Region-CLIP~\cite{region-clip}, which pairs image crops with phrase-level text, and more recent advances like FG-CLIP~\cite{xie2025fgclip}, which scales alignment by constructing a corpus of 1.6B long captions and 40M region-level boxes and training RoI features to contrast with phrases.

Moving beyond explicit regions, patch-text alignment methods more abstractly refine these correspondences by grouping vision patches with text tokens—typically using similarity measures or heuristics for averaging \cite{yao2022filip, mukhoti2022pacl, bica2024sparc} or through trainable modules for aggregating patches and text tokens as in \cite{asokan2025}. Notably, DreamLIP~\cite{zheng2024dreamlip} observed that sentences within MLLM-generated long captions often describe only partial aspects of an image—such as a specific object or scene. Rather than relying on region-level supervision or explicit token grouping, it decomposes each long caption into sub-captions to create multiple positive text pairs, aligning them globally with a multi-positive contrastive loss and locally via cosine-similarity–based patch pooling.
FLAIR~\cite{xiao2024flair} advances this idea by replacing the cosine-based pooling with text-conditioned attention pooling that extracts patch features specific to a given query. 

In this work, we adapt FLAIR's attention pooling to select visual features for a \textit{hierarchy of textual queries}. Specifically, $\beta$-CLIP decomposes long captions into constituent sentences and phrases, replacing standard pooling with a shallow Transformer block modified for Cross-Attention Pooling. This refines query--patch interactions to improve fine-grained selection while retaining broader context. By densely probing visual features at multiple granularities, our method concentrates attention on fine-grained details where global representations exhibit poor separability. Unlike FLAIR, which relies on text-conditioned pooling at both training and inference, $\beta$-CLIP uses it only during training and reduces to standard CLIP at inference, preserving its caching efficiency. However, training with multi-granular queries poses unique challenges for contrastive learning due to semantic overlap across granularities. Unlike traditional region alignment—where local regions explicitly align with queries—our pooled features are inherently contextualized within global image semantics, as cross-attention preserves contributions from contextual patch tokens.

To handle overlapping semantics, we propose a $\beta$-contextualized contrastive alignment loss ($\beta$-CAL): a parameterized contrastive objective that treats all feature pairs from the same image as positives. The objective is controlled by a weighting factor $\beta \in [0,1]$, that interpolates between strict and relaxed positive alignments across two contrastive variants—a soft Cross-Entropy (CE) form or a hard Binary Cross-Entropy (BCE) form. Specifically, at $\beta=0$, it enforces strict self-matching (treating only the most precise intra-image pair as a full positive) to prioritize sharp, fine-grained alignments; at higher $\beta$ values (approaching 1), it promotes more uniform competition among all these intra-image positives—tuning soft probability targets in the CE variant or $\beta$-weighted gradients in the BCE variant—to better integrate contextual information without diluting specificity.

\noindent Our primary contributions are summarized as follows:
\begin{itemize}
\item We propose $\beta$-CLIP, a multi-granular text-conditioned contrastive learning framework that densely aligns image representations with hierarchical text descriptions, demonstrating the effectiveness of probing visual features at multiple granularities for fine-grained understanding.
\item We introduce a $\beta$-weighted contrastive objective ($\beta$-CAL) to address overlapping semantics in text-conditioned patch pooling. This objective supports both smooth-target cross-entropy and hard-target binary cross-entropy losses, enabling fine-grained learning from long captions.
\item We validate $\beta$-CLIP by fine-tuning CLIP on long-text and image pairs from ShareGPT4V, yielding significant improvements on the challenging fine-grained and long-text retrieval benchmarks FG-OVD, DCI, and Urban1K, without relying on hard negatives.
\item We identify a trade-off between specificity and contextualization when training on long captions, which can lead to overfitting in long-text retrieval tasks. We demonstrate that this issue affects different types of losses differently and can be navigated using the $\beta$-CAL loss.
\end{itemize}

%% file: figs/visualizations.tex
\begin{figure}[ht]
    \centering
    \renewcommand{\arraystretch}{0.0}
    \setlength{\tabcolsep}{0pt} 
    \begin{tabular}{@{}c@{}c@{}c@{}c@{}c@{}}
    \rotatebox{90}{\parbox{2.2cm}{\centering \scriptsize Original Image}} &
    \includegraphics[width=0.12\textwidth]{figs/visualizations_final/originals/sa_1544018_Despite_the_background_the_colorful_tuk-tuks_and_chatting_locals_stand_out_in_this_image._pos.png} &
    \includegraphics[width=0.12\textwidth]{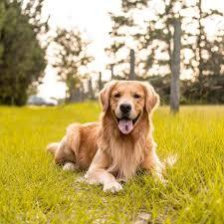} &
    \includegraphics[width=0.12\textwidth]{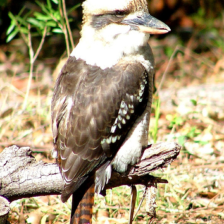} &
    \includegraphics[width=0.12\textwidth]{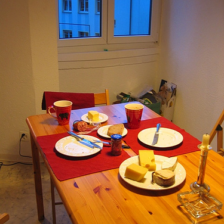}
    \\[-0.25pt]
    \rotatebox{90}{\parbox{2.2cm}{\centering \scriptsize CLIP ViT-B/16}} &
    \includegraphics[width=0.12\textwidth]{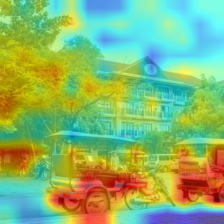} &
    \includegraphics[width=0.12\textwidth]{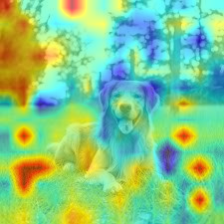} &
    \includegraphics[width=0.12\textwidth]{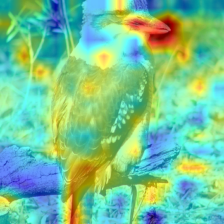} &
    \includegraphics[width=0.12\textwidth]{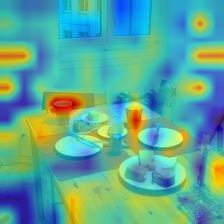}
    \\[-3pt]
    \rotatebox{90}{\parbox{2.2cm}{\centering \scriptsize $\beta$-CLIP-CE, K=6}} &
    \includegraphics[width=0.12\textwidth]{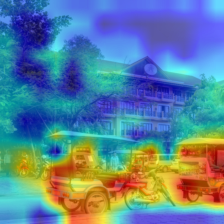} &
    \includegraphics[width=0.12\textwidth]{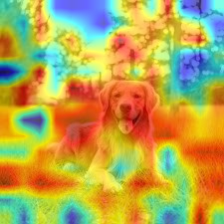} &
    \includegraphics[width=0.12\textwidth]{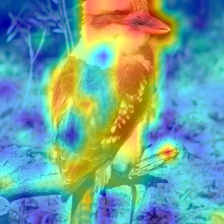} &
    \includegraphics[width=0.12\textwidth]{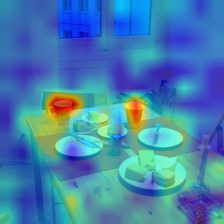}
    \\[-3pt]
    \rotatebox{90}{\parbox{2.2cm}{\centering \scriptsize $\beta$-CLIP-CE, K=36}} &
    \includegraphics[width=0.12\textwidth]{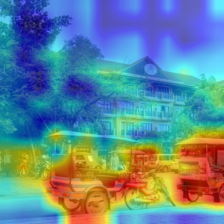} &
    \includegraphics[width=0.12\textwidth]{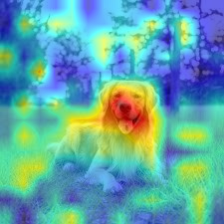} &
    \includegraphics[width=0.12\textwidth]{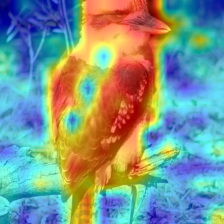} &
    \includegraphics[width=0.12\textwidth]{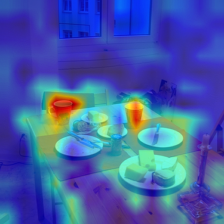}
    \\[-3pt]
    \\[6pt]
    &
    {\tiny \parbox{0.125\textwidth}{\centering \texttt{Despite the background, the colorful tuk-tuks and chatting locals stand out in this image.}}} &
    {\scriptsize \parbox{0.125\textwidth}{\centering \texttt{Nose}}} &
    {\scriptsize \parbox{0.125\textwidth}{\centering \texttt{A closeup of a bird with its wings closed.}}} &
    {\scriptsize \parbox{0.125\textwidth}{\centering \texttt{Cups of coffee}}}
    \\[-2pt]
    \end{tabular}
    \caption{\textbf{Heatmaps of patch-text logit similarities.} Compared to pretrained CLIP's diffuse global activations, training with 5-sentences ($K=6$) enhances localized peaks for holistic scenes (e.g., tuk-tuks) but shows less precise localization for granular details (e.g., dog's nose); using an additional 30-concept ($K=36$) yields spatially precise, high-contrast activations for the semantic features (e.g., nose, coffee cups). BCE variants in Supplementary Material.}
    \label{fig:visualization}
\end{figure}

%% file: sec/2_relatedwork.tex
\section{Related Work}

\subsection{Fine-Grained Vision-language Alignment}
CLIP’s single global image embedding has motivated many approaches to incorporate fine-grained vision-text alignment in diverse ways. For explicit region-text alignment, RegionCLIP~\cite{region-clip} mines pseudo region–caption pairs for regional contrastive learning, while FineCLIP~\cite{jing2024fineclip} combines object crops with VLM-generated region captions and self-distillation to supervise patch-level features. FG-CLIP~\cite{xie2025fgclip} scales this paradigm with tens of millions of region captions, long descriptions, and hard compositional negatives. Other methods avoid explicit bounding boxes entirely: FILIP~\cite{yao2022filip} and PACL~\cite{mukhoti2022pacl} introduce token-wise or patch-based alignment, while SPARC~\cite{bica2024sparc} learns sparse groupings of patches for specific tokens. Recently, \cite{qiu2025refiningclip} showed that naively projecting all patch tokens into text space can collapse spatial correlations despite strong retrieval performance. Moving beyond such projections, recent works have shifted toward text-conditioned feature selection, where the text query dynamically selects relevant visual features. FLAIR~\cite{xiao2024flair} applies text-conditioned cross-attention to pool specific patch features, while SmartCLIP~\cite{xie2025smartclip} learns a binary mask over visual embeddings to select feature dimensions most relevant to the text. \cite{bianchi2024clip} provides a complementary view of CLIP’s fine-grained limitations, showing that while attributes such as color, material, and patterns are present in its embeddings, contrastive pretraining biases the latent space toward coarse, category-level semantics. This suggests that CLIP’s shortcomings stem not from a lack of localization alone, but from its latent representation, motivating methods that better disentangle fine-grained attributes within the shared image–text space.

\subsection{Long Captions in CLIP} 
CLIP is pre-trained with a 77-token text window, which limits its use with long descriptions. \cite{urbanek2024dci} show that CLIP-based models underutilize rich visual detail and introduce DCI—human-annotated, mask-aligned captions that are $\approx$ 1k words in length—along with a summarized variant, sDCI, that fits within 77 tokens. Fine-tuning CLIP on the small sDCI set (with LLM-generated summaries and negatives) yields improvements on fine-grained discrimination. Similarly efforts have been made by DOCCI \cite{onoe2024docci} to increase fine-grained details in captions and DreamLIP \cite{zheng2024dreamlip} and ShareGPT4V \cite{chen2024sharegpt4v} with synthetically generated long captions. Using ShareGPT4V captions, LongCLIP \cite{zhang2024longclip} shows that by fixing the first 20 positional embeddings and 4x interpolating the rest, CLIP can be fine-tuned to handle longer captions (up to 248 tokens), partially mitigating the context-length bottleneck; this approach is being increasingly adopted \cite{xie2025fgclip,xie2025smartclip,asokan2025}, as do we. TULIP \cite{najdenkoska2025tulip} replaces absolute encodings with relative ones and uses distillation plus long-caption finetuning to finetune CLIP to arbitrary token lengths. Interestingly, \cite{wu2024lotlip} observe that training with long captions boosts long-text retrieval but degrades short-text retrieval; they subsequently propose using learnable tokens in the text encoder to aggregate diverse features via attention masks, which helps recover short-text performance. In our work, we also observe this phenomena in the CE variants, and observe that the BCE variants can help mitigate this.

%% file: sec/3_method.tex
\section{Methodology}
\begin{figure*}[t]
  \centering
\includegraphics[width=\textwidth]{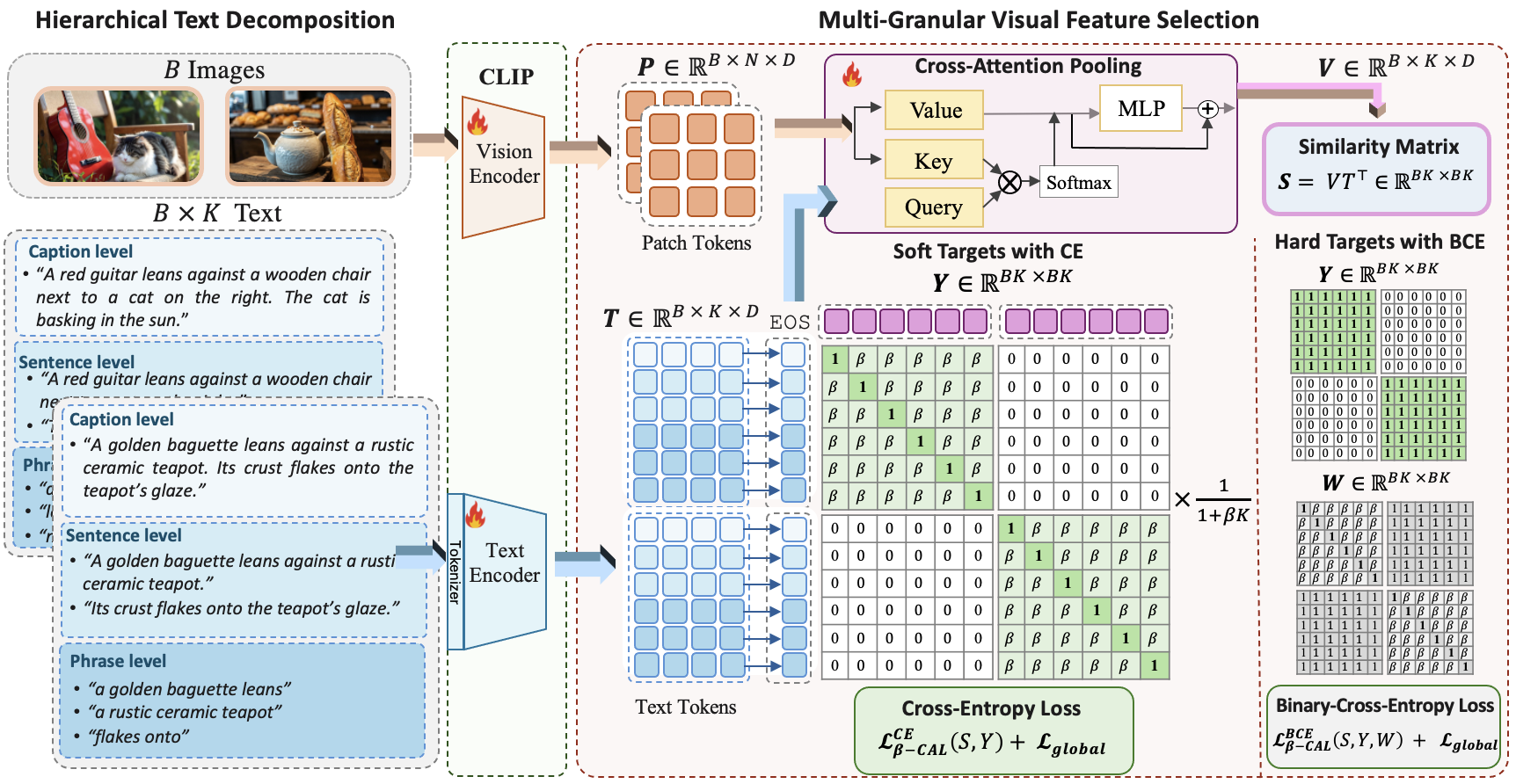}
    \caption{\textbf{Overview of the $\beta$-CLIP architecture} for a batch size of $B=2$ and number of text queries $K=6$ for illustrative purposes. For each image, the text is decomposed into $K$ multi-granular inputs and independently encoded using the CLIP vision and text encoders respectively. The patch tokens $\mathbf{P}$ (\includegraphics[height=0.9em]{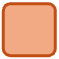}\hspace{0em}) and \texttt{EOS} tokens $\mathbf{T}$ (\includegraphics[height=0.9em] {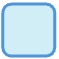}\hspace{0em}) serve as the keys/values and queries to compute the multi-granular, text-conditioned visual embeddings $\mathbf{V}$ (\includegraphics[height=0.9em]{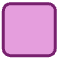}\hspace{0em}) via the \textbf{cross-attention pooling block}. We compute the similarities between the flattened text tokens $\mathbf{T}$ and the flattened text-conditioned visual embeddings $\mathbf{V}$ and symmetrically apply $\beta$-CAL using either normalized \textbf{soft targets with cross-entropy (CE)} or \textbf{hard targets with binary cross-entropy (BCE)}. For, BCE, the $\beta$-weights are applied to loss as described in Eq.~(\ref{eq:BCE}). 0.We omit the visualizations for the \texttt{CLS} tokens and the details of the global loss for the \texttt{CLS} tokens and caption-level text \texttt{EOS} tokens.}
  \label{fig:fig1}
\end{figure*}

Our approach extends CLIP for fine-grained, multi-granular alignment by decomposing each caption into sentence- and phrase-level queries. These queries guide text-conditioned attention pooling—implemented with a shallow modified Transformer block—to extract scale-specific features. We align these features using the $\beta$-contextualized contrastive alignment loss ($\beta$-CAL), which balances fine-grained precision with contextual integration. Figure~\ref{fig:fig1} provides an overview.


\subsection{Hierarchical Text Decomposition}
Given an image–caption pair $(I, C)$, we decompose the caption into three semantic scales.

\noindent\textbf{Caption level:} The full caption provides global context:
$$\mathbf{t}_{\text{cap}} = f_{\text{text}}(C) \in \mathbb{R}^D,$$
where $f_{\text{text}}$ is the CLIP text encoder.

\noindent\textbf{Sentence-Level}: We segment the caption into $K_{\text{sent}}$ individual sentences for coarse-grained semantics:
$$\{\mathbf{t}_{\text{sent}}^i\}_{i=1}^{K_{\text{sent}}} = \{f_{\text{text}}(s_i) : s_i \in \text{split}(C)\}.$$

\noindent\textbf{Phrase-Level}: We extract $K_{\text{phrase}}$ key concepts (noun phrases and verb phrases) using dependency parsing (via spaCy \cite{honnibal2020spacy}) for localized/fine-grained semantics:
$$\{\mathbf{t}_{\text{phrase}}^j\}_{j=1}^{K_{\text{phrase}}} = \{f_{\text{text}}(p_j) : p_j \in \text{extract}(C)\}.$$
This yields $K = 1 + K_{\text{sent}} + K_{\text{phrase}}$ text embeddings per image, forming the multi-scale text representation $\mathbf{T} = [\mathbf{t}_1, \ldots, \mathbf{t}_K] \in \mathbb{R}^{K \times D}$. Since the CLIP text encoder is trained with a causal mask, we need to extract the \texttt{EOS} token for each text query rather than simply reusing the corresponding text tokens from the full caption.

\subsection{Multi-Granularity Visual Feature Selection}
For the input image $I$, we extract patch embeddings using the CLIP vision transformer using the approach from \cite{mask-clip}, skipping the $Q-K$ mixing in the final attention block for the patch tokens of the ViT to preserve localized features:
$$ 
[\mathbf{v}_{\text{CLS}}; \mathbf{P}] = f_{\text{vision}}(I) \in \mathbb{R}^{(N + 1) \times D},$$
where $\mathbf{P} = [\mathbf{p}_1, \ldots, \mathbf{p}_N]$ are the patch tokens and $N$ is the number of visual patches ($N=196$ for ViT-B/16) and $\mathbf{v}_{\text{CLS}} \in \mathbb{R}^{D}$ is the global image token. We then generate $K$ text-conditioned image embeddings via cross-attention pooling. For each text query $\mathbf{t}_k \in \mathbf{T}$, we compute attention weights over the patches:
\begin{equation}
\begin{aligned}
\mathbf{Q}_k &= \mathbf{t}_k \mathbf{W}_Q, \quad \mathbf{K} = \mathbf{P} \mathbf{W}_K, \quad \mathbf{V} = \mathbf{P} \mathbf{W}_{V}, \\
\alpha_k &= \mathrm{softmax}\!\left(\frac{\mathbf{Q}_k \mathbf{K}^\top}{\sqrt{D/h}}\right),\quad
\mathbf{v}_k = \alpha_k \mathbf{V}.
\end{aligned}
\end{equation}
where $\mathbf{W}_Q, \mathbf{W}_K, \mathbf{W}_V \in \mathbb{R}^{D \times D}$ are learned projections, $h=8$ is the number of attention heads, and $D/h$ is the head dimension. The Transformer block is modified to skip the first residual connection around the multi-head self-attention layer. Instead, after computing the attention-weighted values, the model directly normalizes and applies a 2-layered MLP to each query-specific value vector $\mathbf{v}_k$, followed by a residual connection. This produces query-specific visual representations $\mathbf{V} = [\mathbf{v}_1, \ldots, \mathbf{v}_K] \in \mathbb{R}^{K \times D}$, where each $\mathbf{v}_k$ attends to visual regions relevant to $\mathbf{t}_k$. 
\subsection{$\beta$-Contextualized Contrastive Alignment Loss}
Since the hierarchical features introduce semantic overlaps, phrase features $\mathbf{v}_j^{phrase}$ may subsume parts of sentence features $\mathbf{v}_i^{sent}$. Given this contextualization, we treat \emph{all} intra-image feature pairs as positives with tunable overlap via $\beta \in [0,1]$. For a batch of $B$ images, both variants construct a flattened similarity matrix $\mathbf{S} \in \mathbb{R}^{BK \times BK}$ where $S_{ij} = \mathbf{v}_i^\top \mathbf{t}_j / \tau$, and $\tau$ is a learned temperature. We then apply $\beta$ differently for each variant: in the soft Cross-Entropy (CE) form, it interpolates probabilistic targets; in the hard Binary Cross-Entropy (BCE) form, it modulates gradient weights on binary positives, as detailed below.

\subsubsection{Soft Targets with Cross-Entropy}
\label{sec:soft_targetss_with_ce}
To form the training targets, we first define the target weights $w_{ij}$, which represent the desired strength of alignment between features $i$ and $j$. These weights label exact matches as positives, related intra-image pairs as tunable soft positives modulated by $\beta$, and cross-image pairs as negatives:
\begin{equation}
\resizebox{0.9\linewidth}{!}{$
w_{ij}=
\begin{cases}
1 & \text{if } i=j\;(\text{positive})\\
\beta  & \text{if } \lfloor i/K \rfloor=\lfloor j/K \rfloor,\; i\neq j\;(\text{intra-image positives})\\
0      & \text{otherwise}\;(\text{cross-image negatives})
\end{cases}
$}
\label{eq:weights}
\end{equation}
and row-normalize them to get a soft distribution
$$p_{ij} = \frac{w_{ij}}{\sum_{l=1}^{BK} w_{il}}.$$ We use these weights in lieu of the labels.
Per standard Cross-Entropy, we apply \emph{softmax} to the similarities $\mathbf{S}$ to yield predicted probability distributions $q_{ij}^{\mathrm{v2t}}$ and $q_{ji}^{\mathrm{t2v}}$.
The $\beta$-CAL loss is computed as the symmetric cross-entropy \cite{oord2019infonce}:
\begin{equation}
\resizebox{0.5\columnwidth}{!}{$
\begin{aligned}
\mathcal{L}^{\mathrm{CE}}_{\mathrm{v2t}} &= -\frac{1}{BK}\sum_{i,j} p_{ij}\log q^{\mathrm{v2t}}_{ij},\\
\mathcal{L}^{\mathrm{CE}}_{\mathrm{t2v}} &= -\frac{1}{BK}\sum_{i,j} p_{ij}\log q^{\mathrm{t2v}}_{ji},\\
\mathcal{L}^{\mathrm{CE}}_{\text{$\beta$-CAL}} &= \tfrac{1}{2}\big(\mathcal{L}^{\mathrm{CE}}_{\mathrm{v2t}}+\mathcal{L}^{\mathrm{CE}}_{\mathrm{t2v}}\big)
\end{aligned}
$}
\end{equation}
These soft interpolated weights promotes exact self-alignment (via the diagonal $w_{ij}=1$) while controlling the alignment of contextual information from related intra-image features (via $\beta$). When $\beta=0$, only exact matches contribute; as $\beta$ increases to 1, the full intra-image set acts as positives. The softmax step ensures the predictions form a valid probability distribution over all possible pairs.
The complete objective combines $\beta$-CAL with standard CLIP loss where the $\mathcal{L}_{\text{global}}$ is applied to the similarities between the global image tokens and the full caption tokens respectively:
$$\mathcal{L}_{\text{total}} = \mathcal{L}_{\text{$\beta$-CAL}} + \mathcal{L}_{\text{global}}.$$

\subsubsection{Hard Targets with Binary Cross-Entropy}
In the Binary Cross Entropy formulation, the targets $\mathbf{Y}=\{y_{i,j}\}$ remain strictly binary such that \emph{all} intra-image pairs are positives:

$$y_{ij} = \begin{cases}
1 & \text{if } \lfloor i/K \rfloor = \lfloor j/K \rfloor \\
0 & \text{otherwise},
\end{cases}$$
The similarities $\mathbf{S}$ are passed through a \emph{sigmoid} activation to yield independent binary probabilities $q^{\mathrm{v2t}}_{ij} = \sigma(S_{ij})$ and $q^{\mathrm{t2v}}_{ji} = \sigma(S_{ji})$~\cite{zhai2023siglip}. The weights $\mathbf{W}=\{w_{i,j}\}$  for modulating the loss are computed as:
\begin{equation}
\resizebox{0.9\linewidth}{!}{$
w_{ij}=
\begin{cases}
1 & \text{if } i=j\;(\text{positive})\\
\beta  & \text{if } \lfloor i/K \rfloor=\lfloor j/K \rfloor,\; i\neq j\;(\text{intra-image positives})\\
1      & \text{otherwise}\;(\text{cross-image negatives})
\end{cases}
$}
\label{eq:weights}
\end{equation}

We then compute the symmetric BCE loss as:
\begin{equation}
\resizebox{0.8\columnwidth}{!}{$
\begin{gathered}
\mathcal{L}^{\mathrm{BCE}}_{\mathrm{v2t}}
= -\frac{1}{BK}\sum_{i,j}w_{ij}\!\Big[y_{ij}\log q^{\mathrm{v2t}}_{ij}+(1-y_{ij})\log\!\big(1-q^{\mathrm{v2t}}_{ij}\big)\Big],\\
\mathcal{L}^{\mathrm{BCE}}_{\mathrm{t2v}}
= -\frac{1}{BK}\sum_{i,j}w_{ij}\!\Big[y_{ji}\log q^{\mathrm{t2v}}_{ji}+(1-y_{ji})\log\!\big(1-q^{\mathrm{t2v}}_{ji}\big)\Big],\\
\mathcal{L}^{\mathrm{BCE}}_{\text{$\beta$-CAL}}
= \tfrac{1}{2}\big(\mathcal{L}^{\mathrm{BCE}}_{\mathrm{v2t}}+\mathcal{L}^{\mathrm{BCE}}_{\mathrm{t2v}}\big).
\end{gathered}
$}
\label{eq:BCE}
\end{equation}
The weights $w_{ij}$ (from Eq.~\ref{eq:weights}) scale each pair's contribution.
In this way, this BCE formulation assigns binary positives to all intra-image pairs, but modulates their gradient contributions via $ \beta $, downweighting off-diagonal terms when $\beta<1$. By scaling the influence of these contextual positives, it provides an alternative way to navigate the precision-generalization trade-off. 

\subsection{The Role of $\beta$}
\label{sec:trade-off}
The parameter $\beta$ controls the balance between query-specific precision and intra-image contextualization:

\begin{itemize}
    \item As $\beta \to 0$: emphasizes self-matching (only the diagonal target is 1), but overfits by collapsing multi-scale representations, ignoring hierarchical structure.
    \item As $\beta \to 1$: uniformly distributes the targets for the intra-image positives, encouraging cross-scale consistency, but potentially diluting query-specific signals.
\end{itemize}

This behavior is characterized by the diagonal fraction $f_{\text{diag}} = 1 / \big(1 + (K-1)\beta\big)$, which determines how much target mass is assigned to the self-match relative to other intra-image positives. Small $\beta$ yields a high $f_{\text{diag}}$, producing highly selective, text-specific image features, while larger $\beta$ lowers $f_{\text{diag}}$ and encourages stronger cross-scale consistency.


%% file: sec/4_experiments.tex
\section{Experiments}
\label{sec:experiments}
We evaluate $\beta$-CLIP on fine-grained, long-text, and coarse-grained retrieval tasks. Models are fine-tuned on a safety-filtered subset of ShareGPT4V-1.24M\footnote{We remove $\approx$ 2K image–text pairs containing abusive content using Detoxify \cite{detoxify} for text and FalconsAI \cite{falconsai} for images (full filtering details in Supplementary~\ref{supp:filtering}).} using AdamW with learning rates of $10^{-5}$ for the pre-trained parameters and $10^{-3}$ for the added Transformer block, an effective batch size of 2048 (per-GPU batch size of 64, 4 A100 GPUs, 8 gradient accumulation steps), and 10 training epochs. We note that because each image is paired with $K$ captions of varying granularity, the resulting similarity matrix is of size $BK\times BK$, meaning each image is compared against significantly more negatives than the nominal batch size suggests. See supplementary for more details. 

\subsection{Fine-Grained Retrieval}
\input{tables/fgovd+long_retrieval}
FG-OVD~\cite{bianchi2023devil-cdf} evaluates fine-grained alignment using region-level queries derived from bounding-box annotations. Each region is paired with a positive description and ten hard negatives as semantic distractors. Performance is measured across four difficulty splits—Trivial, Easy, Medium, and Hard. Following~\cite{xie2025fgclip}, we extract region features using RoI pooling over patch embeddings and compute alignment via cosine similarity to text queries. Baselines highlight the task’s difficulty: CLIP achieves 12.0\% on the Hard split, while region-supervised methods such as FG-CLIP reach 46.1\% when trained with hard-negatives.

As shown in Table \ref{tab:fgovd_long_retrieval}, the $\beta$-CLIP variants outperform CLIP and other baselines on fine-grained retrieval benchmarks. With the Cross-Entropy (CE) objective, accuracy increases sharply from the CLIP baseline across all FG-OVD splits, surpassing the region-aligned Fine-CLIP by +4.1, +5.6, +10.0, and +8.4 points on the Hard, Medium, Easy, and Trivial splits, respectively. Notably, $\beta$-CLIP outperforms all methods trained on the same ShareGPT4V-1M data—such as Long-CLIP and Smart-CLIP—demonstrating that data alone does not account for the performance increase. We further observe the effectiveness of our multi-granular text-conditioned approach when compared to the region-aligned FG-CLIP$^{\dagger\dagger}$. Despite using significantly less training data (1.2M vs. 1.6B+12M+50M region proposals) and training without hard negatives, the $\beta$ contextualized alignment allows us to recover 55\% of the performance gap between standard CLIP and FG-CLIP—calculated as the ratio of our absolute gain over CLIP to the FG-CLIPs absolute gain over CLIP.

\subsection{Long-Text Retrieval}
We evaluate zero-shot retrieval on extended captions using the DCI, SV-1k, and Urban-1k benchmarks, where long descriptions challenge CLIP’s 77-token limit and its global alignment objective. DCI \cite{urbanek2024dci} provides dense, region-level narratives spanning multiple entities per image. SV-1k~\cite{chen2024sharegpt4v} contains $1K$ validation ShareGPT4V captions and serves as an in-distribution evaluation. Urban-1k~\cite{zhang2024longclip} comprises visually similar urban scenes from Visual Genome~\cite{krishna2016visualgenome}, each paired with a GPT-4V–generated caption averaging 107 words. As shown in table \ref{tab:fgovd_long_retrieval}, $\beta$-CLIP (BCE) establishes a new state-of-the-art on the Urban-1k benchmark, achieving \textbf{91.8\%} (T2I) and \textbf{92.3\%} (I2T). Notably, it outperforms models specifically designed for long-text understanding trained on the same data, including Smart-CLIP (87.4\% T2I) and Long-CLIP (79.5\% T2I). On the DCI benchmark, which requires understanding multiple entities and relationships, $\beta$-CLIP achieves 65.1\% R@1, surpassing Long-CLIP by a wide margin (+7.7\%) and performing comparably to FLAIR and SmartCLIP.

\subsection{Coarse-Grained Retrieval}
\input{tables/short_fg_retrieval}

Standard coarse-grained retrieval is evaluated on short-caption datasets (MSCOCO and Flickr30k) to assess whether multi-granular training preserves CLIP’s global alignment behavior. Across both datasets, methods that rely on sigmoid-based objectives—notably SigLIP and FLAIR—consistently outperform CLIP and other CE-based variants, suggesting that optimizing for independent binary matching helps with coarse-grained retrieval. 

As shown in Table \ref{tab:retrieval_full}, the CE-based $\beta$-CLIP model performs comparably to the CLIP baseline. In contrast, the BCE variant of $\beta$-CLIP consistently achieves stronger performance on these tasks, surpassing both the CE variant and the CLIP baseline. Notably, whereas prior work has shown that naively fine-tuning on detailed captions can degrade short-text alignment \cite{zhang2024longclip, wu2024lotlip}, our BCE variant maintains absolute gains over the CLIP baseline on both MSCOCO (+6.7 T2I and +1.5 I2T R@1) and Flickr30k (+7.7 T2I and +1.5 I2T R@1), comparable to the improvements achieved by dedicated long-caption methods such as LongCLIP and SmartCLIP.

%% file: tables/fgovd+long_retrieval.tex
\begin{table*}[!htbp]
  \centering\tiny 
  \SetTblrInner{rowsep=0.25pt, colsep=1pt} 
  \resizebox{0.75\linewidth}{!}{ 
  \begin{tblr}{
      colspec = {l Q[l,wd=1.7cm] *{4}{c} !{\vrule width 0.1pt} *{6}{c}},
      rows = {m},
      row{1-3} = {font=\bfseries\fontsize{4pt}{4.5pt}\selectfont},
      column{1} = {halign=l},
      column{2} = {halign=c, font=\fontsize{4pt}{4pt}\selectfont},
      column{3-Z} = {halign=c},
      hline{1,Z} = {1pt},
      hline{4} = {0.2pt},
      hline{3}={7-8}{leftpos=-1,rightpos=-1,endpos},
      hline{3}={9-10}{leftpos=-1,rightpos=-1,endpos},
      hline{3}={11-12}{leftpos=-1,rightpos=-1,endpos},
      cell{2}{7} = {c=2}{c},
      cell{2}{9} = {c=2}{c},
      cell{2}{11} = {c=2}{c},
  }
  \textbf{Method} & \textbf{Data} & \SetCell[c=4]{c} \textbf{Fine-Grained Retrieval} & & & & \SetCell[c=6]{c} \textbf{Long-Text Retrieval} & & & & & \\
   & & \SetCell[c=4]{c} \textbf{FG-OVD} & & & & \textbf{DCI} & & \textbf{SV-1k} & & \textbf{Urban-1k} & \\
   \emph{ViT-B/16} & &\textbf{Hard} & \textbf{Medium} & \textbf{Easy} & \textbf{Trivial} & \textbf{T2I} & \textbf{I2T} & \textbf{T2I} & \textbf{I2T} & \textbf{T2I} & \textbf{I2T} \\
  CLIP & WIT-400M & 12.0 & 23.1 & 22.2 & 58.5 & 43.0 & 45.5 & 79.6 & 78.2 & 53.2 & 67.5 \\
  EVA-CLIP & Merged-2B\textsuperscript{\textdagger} & 14.0 & 30.1 & 29.4 & 58.3 & 41.2 & 41.9 & 85.5 & 90.5 & - & - \\
  SigLIP & WebLI-1B & - & - & - & - & 56.2 & 57.7 & 83.4 & 85.8 & 62.1 & 62.7 \\
  Fine-CLIP & {CC2.5M + 10.4M reg.} & \underline{26.8} & \underline{49.8} & \underline{50.4} & \underline{71.9} & 34.4 & 35.5 & 73.3 & 70.6 & - & - \\
  LoTLIP & Mixed-100M\textsuperscript{\textsection} & - & - & - & - & 61.0 & 62.1 & 86.8 & 95.5 & - & - \\
  FLAIR* & DreamLIP 30M & 13.3 & 27.5 & 28.6 & 79.3 & \textbf{66.2} & 61.3 & 98.0 & 98.5 & 87.7 & 83.6 \\
  FG-CLIP \textsuperscript{\textdaggerdbl} & \SetCell[r=1]{c}{WIT-400M + LAION-1.6B + GRIT-12M + 40M reg.} & 24.5 & 47.1 & 49.5 & - & - & - & - & - & - & - \\
  \textcolor{gray}{FG-CLIP \textsuperscript{\textdagger\textdagger}} & \textcolor{gray}{+10M hard negatives} & \textcolor{gray}{46.1} & \textcolor{gray}{66.6} & \textcolor{gray}{ 68.7} & \textcolor{gray}{83.4} & \textcolor{gray}{60.6} & \textcolor{gray}{61.8} & \textcolor{gray}{94.9} & \textcolor{gray}{96.7} & \textcolor{gray}{89.9} & \textcolor{gray}{93.0} \\
  \hline[0.2pt]
  \SetCell[c=12]{l}{\emph{Methods fine-tuning CLIP on WIT-400M + ShareGPT4V-1M }} \\
  \hline[0.2pt]

  Long-CLIP\textsuperscript{*} &  \SetCell[r=4]{c}{WIT-400M\\+ ShareGPT4V-1M} & 9.2 & 18.4 & 16.2 & 51.8 & 57.4 & 51.8 & 93.4 & 94.7 & 79.5 & 78.9 \\
  TULIP & & - & - & - & - & 50.6 & 50.2 & \textbf{98.6} & \underline{98.6} & 86.6 & 88.1  \\
  FineLIP & & - & - & - & - & - & - & 89.3 & 90.7 & 87.4 & \underline{90.0} \\
  Smart-CLIP\textsuperscript{*} &  & 18.9 & 37.0 & 37.9  & 70.1 & 64.5 & \textbf{65.3} & \underline{98.1} & \textbf{98.7} & 87.4 & \underline{90.0} \\
  \hline[0.2pt]
  \SetCell[c=12]{l}{\emph{Ours: $\beta$-CLIP}} \\
  \hline[0.2pt]
  \text{\tiny K=36, $\beta$=0.5, CE} & \SetCell[r=2]{c}{WIT-400M\\+ ShareGPT4V-1M} & \textbf{30.9} & \textbf{55.4} & \textbf{60.4} & \textbf{80.3} & 59.9 & 58.4 & 93.7 & 94.0 & \underline{89.0} & 88.6 \\
  \text{\tiny K=36, $\beta$=0.5, BCE} & & 20.1 & 38.5 & 34.2 & 71.3 & \underline{65.1} & \underline{63.6} & 94.4 & 94.1 & \textbf{91.8} & \textbf{92.3}  \\
  \end{tblr}}
  \caption{Performance on fine-grained and long-text retrieval tasks. \textsuperscript{\textdaggerdbl} denotes FG-CLIP trained \textbf{without} 10M hard negatives while  \textsuperscript{\textdagger\textdagger} denotes FG-CLIP trained \textbf{with 10M hard negatives}. We \textcolor{gray}{gray} out FG-CLIP for fair comparison with our method, which does not use Hard Negatives during training.
    \textsuperscript{\textdagger} marks models trained on a dataset combining 1.6B LAION-2B and 0.4B COYO-700M samples. 
    \textsection refers to a 100M-image recaptioned dataset (CC3M, CC12M, YFCC15M, LAION, COYO). 
    “reg.” denotes the use of additional region-level supervision. \textsuperscript{*} indicates models we evaluate (FG-OVD and DCI for SmartCLIP and LongCLIP and FG-OVD for FLAIR).}

  \label{tab:fgovd_long_retrieval}
\end{table*}

%% file: tables/short_fg_retrieval.tex
\begin{table*}[htbp]
  \centering\tiny
  \SetTblrInner{rowsep=0.25pt}
  \SetTblrInner{colsep=2pt}
  \resizebox{0.75\linewidth}{!}{
    \begin{tblr}{
      colspec = {l l *{4}{c} !{\vrule width 0.1pt} *{4}{c}},
      cells={halign=c,valign=m},
      column{1}={halign=l},
      column{2} = {halign=c},
      hline{1,Z}={1pt},
      hline{5}={},
      hline{3}={3-6}{leftpos=-1, rightpos=-1, endpos},
      hline{3}={7-10}{leftpos=-1, rightpos=-1, endpos},
    }
      & & \SetCell[c=8]{c}{\bf Coarse-Grained Retrieval} \\
      Method & Data
      & \SetCell[c=4]{c}{MSCOCO} & & &
      & \SetCell[c=4]{c}{Flickr30k} \\
      & &
      \SetCell[c=2]{c}{T2I} & &
      \SetCell[c=2]{c}{I2T} & &
      \SetCell[c=2]{c}{T2I} & &
      \SetCell[c=2]{c}{I2T} & & \\
      \emph{ViT-B/16} & & R@1 & R@5 & R@1 & R@5 & R@1 & R@5 & R@1 & R@5 \\
      CLIP & WIT-400M & 33.1 & 58.4 & 52.5 & 76.7 & 62.1 & 85.6 & 81.9 & 96.2 \\
      SigLIP & WebLI-1B & \underline{47.2} & \underline{72.1} & \underline{65.5} & \underline{86.2} & \underline{75.6} & \underline{92.8} & 89.1 & 98.6 \\
      FineCLIP & \SetCell[r=1]{c}{WIT-400M \\+ CC2.5M + 10.4M reg.} & 40.2 & 66.5 & 54.5 & 78.6 & 67.9 & 89.1 & 82.5 & 96.4 \\
      LoTLIP & Mixed-100M & 38.06 & 63.81 & 59.66 & 81.50 &  65.22 & 87.98 &  86.90 & 97.80 \\
      DreamLIP & \SetCell[r=2]{c}{DreamLIP-30M} & 44.8 & 69.8 & 62.3 & 84.5 & 73.3 & 91.8 & 89.9 & \underline{99.0} \\
      FLAIR & & \textbf{53.3} & \textbf{77.5} & \textbf{68.0} & \textbf{87.8} & \textbf{81.1} & \textbf{94.9} & \textbf{94.7} & \textbf{99.3} \\
      \textcolor{gray}{FG-CLIP} & \SetCell[r=1]{c}{\textcolor{gray}{{WIT-400M + LAION-1.6B}} \\ \textcolor{gray}{+ GRIT-12M + 40M reg} \\ \textcolor{gray}{+ 10M hard negatives}} & \textcolor{gray}{45.4} & - & \textcolor{gray}{64.1} & - & \textcolor{gray}{76.4} & - & \textcolor{gray}{90.7} & - \\
      \hline[0.6pt]
      \SetCell[c=10]{l}{\emph{\tiny{Methods fine-tuning CLIP on WIT-400M + ShareGPT4V-1M}}} \\
      \hline[0.2pt]
      TULIP & \SetCell[r=5]{c}{WIT-400M\\+ ShareGPT4V-1M} & 40.7  & 66.1  & 56.8  & 80.3 & -  & - & -  & -  \\
      LongCLIP\textsuperscript{*}  & & 40.4  & 65.8  & 57.6  & 81.1 & 70.4  & 90.5 & 84.1  & 98.0 \\
      SmartCLIP\textsuperscript{*}  & & 42.4 & 68.2  & 61.9  & 83.3 & 72.5  & 91.4 & \underline{90.1}  & 98.4  \\
      \text{$\beta$-CLIP \tiny K=36, $\beta$=0.5, CE} & & 34.0 & 60.2  & 46.0 & 74.2 & 62.6  & 87.1  & 73.2  & 95.2  \\
      \text{$\beta$-CLIP \tiny K=36, $\beta$=0.5, BCE} & & 39.8 & 66.0 & 54.0 & 79.3 & 69.8 & 90.5 & 83.4 & 97.2 \\
\end{tblr}}
    \caption{
    Standard image--text retrieval with ViT-B/16. $\beta$-CLIP variants largely preserve CLIP's performance on MSCOCO and Flickr30k, preventing degradation on coarse-grained retrieval typically observed when finetuning clip on long-text.\textsuperscript{*}  denotes our evaluation on Flickr30k using \cite{cherti_clip_bench}.}
  \label{tab:retrieval_full}
\end{table*}

%% file: sec/5_ablations.tex
\section{Ablations}
We conduct a comprehensive ablation study to validate the core design choices of $\beta$-CLIP. Specifically, we analyze the impact of the alignment objective, the balancing hyperparameter $\beta$, hierarchy size $K$, and text-conditioning strategies on both fine-grained and long-text retrieval tasks.

\paragraph{Contextualized Contrastive Alignment.}
Table~\ref{tab:baselines} isolates the impact of our proposed objective by comparing standard fine-tuning against $\beta$-CLIP. We apply Symmetric Cross-Entropy (CE) with either a single positive ($K{=}1$) long caption or strictly intra-image positive sentences ($K{=}6$). Direct fine-tuning without text-conditioning demonstrates a limited capacity to leverage multiple positives. In contrast, applying the $\beta$-CAL objective (with $\beta{=}0.5$) increases FG-OVD performance by 7--12 points across the Hard, Medium, and Easy splits compared to the baseline. 
Interestingly, at K=1 $\beta$=0 where no sub-caption hierarchy is used but text-conditioning is still present—CE and BCE perform similarly (Hard: $\approx$22, U-1k: $\approx$88). The distinction emerges once hierarchy is introduced. Although both losses benefit from hierarchical supervision, they interact with it differently. CE improves fine-grained discrimination (Hard: 22
$\rightarrow$29.2 at K=6), whereas BCE favors long-text retrieval (U-1k: 87.6$\rightarrow$91.8, 89.5$\rightarrow$92.0 at K=6).

\input{tables/ablation_baselines}

\paragraph{The Specificity--Contextualization Trade-off.}
The hyperparameter $\beta$ governs the trade-off between diagonal specificity and off-diagonal contextualization discussed in Section~\ref{sec:trade-off}. Tables \ref{tab:vary_beta_k6} and \ref{tab:vary_beta_k36} illustrate this dynamic.
At $\beta=0$, the loss focuses strictly on self-matching. While this yields strong results on long-text retrieval (e.g., CE, $K=36$: 94.4\% SV-1k T2I), it produces highly diverse but weakly contextualized features, leading to overfitting and poor fine-grained performance (e.g., 3.6\% on Hard FG-OVD).
Increasing $\beta$ distributes the target mass across intra-image positives, which significantly benefits fine-grained alignment. For CE, $K=36$, Hard accuracy rises from 3.6\% at $\beta=0$ to a peak of 30.9\% around $\beta=0.5$. These gains incur only minor drops ($<1\%$) in long-text retrieval capabilities. BCE shows smoother FG-OVD gains as in $K=36$, Hard increases steadily from 14.8\% at $\beta=0$ to 20.2\% at $\beta=1$, while long-text accuracy varies by only ~0.5\%.
\input{tables/ablation_varying_beta_k_6}

\input{tables/ablation_varying_beta_k_36}

\paragraph{Increasing Fine-Grained Semantics for Multi-Granular Alignment.}Table~\ref{tab:ablations_k} analyzes the impact of increasing the number of positive phrases. We observe that increasing $K$ from $6$ to $36$ improves performance on the Hard split by $+1.7$ points (reaching $30.9\%$), with even larger gains of $+4.1$ and $+4.0$ on the Medium and Easy splits, respectively. This suggests that phrase-level text-conditioned features effectively localize fine-grained semantics, with performance on fine-grained FG-OVD improving as the number of descriptive phrases increases. Long-text retrieval similarly benefits from larger $K$, with DCI scores increasing by +3.9 (T2I) and +6.9 (I2T). 

The BCE variant, however, shows a different pattern. It generally scores lower on FG-OVD than CE, and performance plateaus or diminishes as $K$ increases (e.g., Easy split accuracy drops to 34.2\% at $K=36$). This pattern reflects BCE's weaker fine-grained separation at $\beta$=0.5. This difference likely stems from the way the BCE loss defines its supervision: all intra-image pairs are treated as independent binary positives, with the loss for off-diagonal pairs downweighted by $\beta$=0.5. This encourages the model to distribute its focus across multiple related matches, thereby effecting its ability to distinguish precise query-image alignments from similar ones. Conversely, the CE variant generates a probability distribution over the logits (via softmax) in addition to normalized target weights, where intra-image positives compete against each other. This has the effect of encouraging the model to assign higher probabilities to exact matches. Despite this, BCE remains highly effective for long-text retrieval, consistently matching or surpassing CE across all $K$ values (e.g., +5.2 to +5.5 gain on DCI T2I).\input{tables/ablation_varying_k_6_11_16_36}

\paragraph{Inference With Text-Conditioned Image Representations.} Table~\ref{tab:tci} compares long-text retrieval performance when using  the text-conditioned image (\texttt{TCI}) representations, by conditioning each image embedding on its ground-truth caption, in lieu of the standard image \texttt{CLS} token. As shown, under the CE loss, \texttt{TCI} yields considerable improvements across both SV-1k and U-1k for $K=6$ and $K=36$, indicating that aligning image features to the specific query text greatly improves long-text matching. However, as $K$, increases, the advantage of the \texttt{TCI} diminishes under binary cross-entropy (BCE) loss, which generally outperforms CE loss when using the \texttt{CLS} token.

\input{tables/ablation_TCI}

\paragraph{Cross-Attention Pooling with Negative Text Queries.} We explore the use of text-conditioned negatives, which introduce additional negative contrasts by sampling texts from specific hierarchy levels across other images in the batch. This approach supplements the positive text-conditioned representations with negative ones. These additional text-conditioned representations are then explicitly treated as negatives within the loss function by augmenting the standard set of negative pairs. As show in Table~\ref{tab:cond_negatives}, at $K=36$, CE does not benefit from this type of conditioning. BCE, however, consistently improves FG-OVD, maintains retrieval, and significantly increases the diversity among the representations.
\input{tables/ablation_varying_cond_negatives}

%% file: tables/ablation_baselines.tex
\begin{table}[!ht]
  \centering\tiny 
  \SetTblrInner{rowsep=0.25pt, colsep=1pt} 
  \resizebox{0.85\columnwidth}{!}{ 
  \begin{tblr}{
      colspec = {l *{4}{c} !{\vrule width 0.1pt} *{4}{c}},
      rows = {m},
      row{1-2} = {},
      column{1} = {halign=l},
      column{2-Z} = {halign=c},
      hline{1,Z} = {0.2pt},
      hline{3} = {0.2pt},
      hline{2}={6-7}{leftpos=-1,rightpos=-1,endpos},
      hline{2}={8-9}{leftpos=-1,rightpos=-1,endpos},
      cell{1}{6} = {c=2}{c},
      cell{1}{8} = {c=2}{c},
  }
   {Method} & \SetCell[c=4]{c} {FG-OVD} & & & & {SV-1k} & & {U-1k} & & \\
   \emph{ViT-B/16} & {Hard} & {Medium} & {Easy} & {Trivial} & T2I & I2T & T2I & I2T & \\
  \SetCell[c=10]{c}{\emph{finetuning with CLIP's global loss}} \\
  \hline[0.2pt]
  Single positive, K=1 & 22.0 & 44.6 & 44.2 & 69.6 & 93.8 & \textbf{94.3} & 88.6 & 88.3  \\
  Multiple positive, K=6 & 19.7 & 40.1 & 40.8 & 67.2 & 93.9 & 93.5 & 89.0 & 87.5  \\

  \hline[0.2pt]
  \SetCell[c=10]{c}{\emph{$\beta$-CLIP  }} \\
  \hline[0.2pt]
   \text{CE \tiny K=1, $\beta$=0}  & 22.0 & 46.3 & 44.5 & 72.7 & 93.9 & 94.2 & 88.4 & 89.2 \\
  \text{BCE \tiny K=1, $\beta$=0} & 21.3 & 44.7 & 43.6 & 70.9 & 94.1 & \textbf{94.3} & 87.6 & 89.5  \\
  \hline[0.2pt]
  CE \text{\tiny K=6, $\beta$=0.5} & \textbf{29.2} &\textbf{ 51.3} & \textbf{56.4} & \textbf{81.5} & 93.5 & 94.0 & 87.9 & 88.4  \\
  BCE \text{\tiny K=6, $\beta$=0.5} & 20.6 & 42.7 & 45.3 & 71.2  & \textbf{94.5} & 94.0 & \textbf{91.8} & \textbf{92.0} \\
  \end{tblr}}
  \caption{Comparison of baseline methods fine-tuning CLIP on long-text without text-conditioning. K=1 includes only the long caption, K=6 includes long-caption and sentence-level captions.}
  \label{tab:baselines}
\end{table}

%% file: tables/ablation_varying_beta_k_6.tex
\begin{table}[t]
  \centering\tiny 
  \SetTblrInner{rowsep=0.25pt, colsep=1pt} 
  \resizebox{0.9\columnwidth}{!}{ 
  \begin{tblr}{
      colspec = {l *{4}{c} | *{4}{c} | c},
      rows = {m},
      row{1-2} = {},
      column{1} = {halign=l},
      column{2-Z} = {halign=c},
      hline{1,Z} = {0.2pt},
      hline{3} = {0.2pt},
      hline{2}={6-7}{leftpos=-1,rightpos=-1,endpos},
      hline{2}={8-9}{leftpos=-1,rightpos=-1,endpos},
      cell{1}{6} = {c=2}{c},
      cell{1}{8} = {c=2}{c},
  }
  {Method} & \SetCell[c=4]{c} {FG-OVD} & & & & {SV-1k} & & {U-1k} & &  \\
   \emph{ViT-B/16} & {Hard} & {Medium} & {Easy} & {Trivial} & {T2I} & {I2T} & {T2I} & {I2T} & {Sim}  \\
  \SetCell[c=10]{c}{\emph{$\beta$-CLIP (CE)}} \\
  \hline[0.2pt]
  \text{\tiny K=6, $\beta$=0} & 5.3 & 10.1 & 9.9 & 32.5 & \textbf{95.0} & \textbf{94.8} & \textbf{91.8} & \textbf{91.0} & 0.17 \\
  \text{\tiny K=6, $\beta$=0.25} & 28.3 & 50.0 & 56.1 & 79.8 & 93.2 & 93.9 & 87.9 & 88.8 & 0.93 \\
  \text{\tiny K=6, $\beta$=0.5} & 29.2 & 51.3 & 56.4 & 81.5 & 93.5 & 94.0 & 87.9 & 88.4 & 0.96 \\
  \text{\tiny K=6, $\beta$=0.75} & \textbf{30.6} & \textbf{52.6} & \textbf{58.5} & \textbf{82.1} & 93.4 & 93.9 & 87.3 & 88.7 & 0.97  \\
  \text{\tiny K=6, $\beta$=1.0} & 29.7 & \textbf{52.6} & 58.3 & 81.4 & 93.5 & 93.9 & 87.1 & 88.3 & 0.98  \\
 
  \hline[0.2pt]
  \SetCell[c=10]{c}{\emph{$\beta$-CLIP (BCE) }} \\
  \hline[0.2pt]
  \text{\tiny K=6, $\beta$=0} & 16.1 & 29.9 & 29.0 & 65.6 & 94.0 & \textbf{94.2} & \textbf{92.1} & \textbf{92.5} & 0.56 \\
    \text{\tiny K=6, $\beta$=0.25} & \textbf{21.7} & 41.8 & 45.3 & 70.0 & 94.1 & \textbf{94.2 }& 91.2 & 92.4 & 0.92 \\
  \text{\tiny K=6, $\beta$=0.5} & 20.6 & \textbf{42.7} & 45.3 & 71.2  & \textbf{94.5} & 94.0 & 91.8 & 92.0 & 0.94 \\
  \text{\tiny K=6, $\beta$=0.75} & 20.6 & 42.4 & \textbf{45.7} & 71.8 & 94.3 & 94.0 & 91.7 & 91.4 & 0.94 \\
  \text{\tiny K=6, $\beta$=1} & 18.8 & 41.7& 44.0 & \textbf{72.6} & 94.2 & \textbf{94.2} & 91.6 & 91.7 & 0.95 \\
  \end{tblr}}
  \caption{Effect of varying $\beta$ at $K=6$. $\beta$ controls the specificity-contextualization tradeoff between diagonal specificity and off-diagonal contextualization.}
  \label{tab:vary_beta_k6}
\end{table}

%% file: tables/ablation_varying_beta_k_36.tex
\begin{table}[t]
  \centering\tiny
  \SetTblrInner{rowsep=0.25pt, colsep=1pt}
  \resizebox{0.9\columnwidth}{!}{
  \begin{tblr}{
      colspec = {l *{4}{c} | *{4}{c} | c},
      rows = {m},
      row{1-2} = {},
      column{1} = {halign=l},
      column{2-Z} = {halign=c},
      hline{1,Z} = {0.2pt},
      hline{3} = {0.2pt},
      hline{2}={6-7}{leftpos=-1,rightpos=-1,endpos},
      hline{2}={8-9}{leftpos=-1,rightpos=-1,endpos},
      cell{1}{6} = {c=2}{c},
      cell{1}{8} = {c=2}{c},
  }
  {Method} & \SetCell[c=4]{c} {FG-OVD} & & & & {SV-1k} & & {U-1k} & &  \\
   \emph{ViT-B/16} & {Hard} & {Medium} & {Easy} & {Trivial} & {T2I} & {I2T} & {T2I} & {I2T} & {Sim}  \\
  \SetCell[c=10]{c}{\emph{$\beta$-CLIP (CE)}} \\
  \hline[0.2pt]
  \text{\tiny K=36, $\beta$=0} & 3.6 & 7.1 & 8.9 & 11.5 & \textbf{94.4} & 93.8 &\textbf{ 90.7} & \textbf{91.4} & 0.14 \\
  \text{\tiny K=36, $\beta$=0.1} & 23.2 & 45.3 & 48.9 & 77.5 & 94.0 & 94.1 & 89.2 & 87.9 & 0.92 \\
  \text{\tiny K=36, $\beta$=0.15} & 27.8 & 51.0 & 56.4 & 79.4 & 93.7 &\textbf{ 94.3} & 88.5 & 87.9 & 0.94 \\
  \text{\tiny K=36, $\beta$=0.5} & \textbf{30.9} & \textbf{55.4} & \textbf{60.4} & 80.3 & 93.7 & 94.0 & 89.0 & 88.6 & 0.98 \\
  \text{\tiny K=36, $\beta$=1} & 30.8 & 54.4 & 59.0 & \textbf{80.6} & 93.4 & 94.1 & 88.7 & 88.7 & 0.98 \\
  \hline[0.2pt]
  \SetCell[c=10]{c}{\emph{$\beta$-CLIP (BCE)}} \\
  \hline[0.2pt]
  \text{\tiny K=36, $\beta$=0} & 14.8 & 30.1 & 31.8 & 64.6 & 94.0 & \textbf{94.3} & 91.9 & 91.6 & 0.66 \\
  \text{\tiny K=36, $\beta$=0.25} & 18.1 & 35.6 & 30.0 & \textbf{72.8} & 94.2 & 94.1 & \textbf{92.1} & 91.9 & 0.96 \\
  \text{\tiny K=36, $\beta$=0.5} & 20.1 & 38.5 & 34.2 & 71.3 & \textbf{94.4 }& 94.1 & 91.8 &\textbf{ 92.3} & 0.98 \\
  \text{\tiny K=36, $\beta$=0.75} & 19.8 & 38.0 & 34.2 & \textbf{72.8} & 94.3 & 93.8 & 91.8 & 91.8 & 0.98 \\
  \text{\tiny K=36, $\beta$=1} & \textbf{20.2} & \textbf{40.4 }& \textbf{36.3} & 70.4 & 94.3 & 93.9 & 92.0 & 91.9 & 0.98 \\
  \end{tblr}}
  \caption{Effect of varying $\beta$ at $K=36$. $\beta$ controls the specificity-contextualization tradeoff between diagonal specificity and off-diagonal contextualization.}
  \label{tab:vary_beta_k36}
\end{table}

%% file: tables/ablation_varying_k_6_11_16_36.tex
\begin{table}[t]
  \centering\tiny 
  \SetTblrInner{rowsep=0.25pt, colsep=1pt} 
  \resizebox{\columnwidth}{!}{ 
  \begin{tblr}{
      colspec = {l *{4}{c} | *{6}{c} },
      rows = {m},
      row{1-2} = {},
      column{1} = {halign=l},
      column{2-Z} = {halign=c},
      hline{1,Z} = {0.2pt}, 
      hline{3} = {0.2pt},
      hline{3}   = {6-11}{leftpos=0,rightpos=0},
      cell{1}{1} = {r=2}{l},                
      cell{1}{2} = {c=4}{c},                     
      cell{1}{6} = {c=2}{c},                     
      cell{1}{8} = {c=2}{c},                     
      cell{1}{10}= {c=2}{c},                     
  }
   {Method}  & \SetCell[c=4]{c} FG-OVD & & & & \SetCell[c=2]{c}DCI & & \SetCell[c=2]{c}SV-1k & & \SetCell[c=2]{c}U-1k & \\
   \emph{ViT-B/16} & Hard & Medium & Easy & Trivial & T2I & I2T & T2I & I2T & T2I & I2T \\
   
  \SetCell[c=10]{c}{\emph{$\beta$-CLIP (CE)}} \\
  \hline[0.2pt]
  \text{\tiny K=6, $\beta$=0.5} & 29.2 & 51.3 & 56.4 & \textbf{81.5} & 56.0 & 51.5 & 93.5 & 94.0 & 87.9 & 88.4 \\
  \text{\tiny K=16, $\beta$=0.5} & 29.2 & 52.2 & 58.6 & 80.8 & 59.7 & 56.5 & 93.3 & \textbf{94.5} & 88.1 & 87.9 \\
  \text{\tiny K=36, $\beta$=0.5} & \textbf{30.9} & \textbf{55.4} & \textbf{60.4} & 80.3 & 59.9 & 58.4 & 93.7 & 94.0 & 89.0 & 88.6 \\
  \hline[0.2pt]
  \SetCell[c=10]{c}{\emph{$\beta$-CLIP (BCE)}} \\
  \hline[0.2pt]
  \text{\tiny K=6, $\beta$=0.5} & 20.6 & 42.7 & 45.3 & 71.2 & 64.9 & 62.7 & \textbf{94.5} & 94.0 & 91.8 & 92.0 \\
  \text{\tiny K=16, $\beta$=0.5} & 20.9 & 42.7 & 41.6 & 71.0 & \textbf{65.2} & 63.2 & 94.0 & 94.2 & \textbf{92.2}& 91.8 \\
  \text{\tiny K=36, $\beta$=0.5} & 20.1 & 38.5 & 34.2 & 71.3 & 65.1 & \textbf{63.6} & 94.4 & 94.1 & 91.8 & \textbf{92.3}  \\
 
  \end{tblr}}
  \caption{Effect of varying $K$. $K=6$: 1 caption + 5 sentences; $K=16$: +10 phrases; $K=36$: +30 phrases.}
  \label{tab:ablations_k}
\end{table}

%% file: tables/ablation_TCI.tex
\begin{table}[t]
  \centering\tiny
  \SetTblrInner{rowsep=0.25pt, colsep=1pt}
  \resizebox{0.65\columnwidth}{!}{
  \begin{tblr}{
    colspec = {l !{\vrule width 0.1pt} l !{\vrule width 0.1pt} *{4}{c}},
    rows = {m},
      row{1-3} = {},
      column{1} = {halign=l},
      column{2-Z} = {halign=c},
      hline{1,Z} = {0.2pt},
      hline{3} = {0.2pt},
  }
   Method & & \SetCell[c=2]{c}SV-1k & & \SetCell[c=2]{c}U-1k & \\
   \emph{ViT-B/16} & Token & T2I & I2T & T2I & I2T \\
  \SetCell[c=6]{c}{\emph{$\beta$-CLIP (CE)}} \\
  \hline[0.2pt]
  \text{\tiny K=6, $\beta$=0.5}& \texttt{CLS} & 93.5 & 94.0 & 87.9 & 88.4 \\
  \text{\tiny K=6, $\beta$=0.5} &  \texttt{TCI} & \textbf{95.0} & \textbf{94.7} & \textbf{99.1}  & \textbf{97.2}  \\
  \hline[0.2pt]
  \text{\tiny K=36, $\beta$=0.5}&  \texttt{CLS} & 93.7 & 94.0 & 89.0 & \textbf{88.6} \\
   \text{\tiny K=36, $\beta$=0.5} & \texttt{TCI} & \textbf{94.7 }& \textbf{94.5 }& \textbf{95.5} & 85.4 \\
  \hline[0.2pt]
  \SetCell[c=6]{c}{\emph{$\beta$-CLIP (BCE)}} \\
  \hline[0.2pt]
  \text{\tiny K=6, $\beta$=0.5} &  \texttt{CLS} & 91.8 & 92.0 & 94.5 & 94.0 \\
  \text{\tiny K=6, $\beta$=0.5} & \texttt{TCI} & \textbf{94.9} & \textbf{94.6} & \textbf{97.0} & \textbf{95.3} \\
  \hline[0.2pt]
  \text{\tiny K=36, $\beta$=0.5} &  \texttt{CLS} & 91.8 & 92.3 & \textbf{94.4} & \textbf{94.1}\\
  \text{\tiny K=36, $\beta$=0.5} & \texttt{TCI} & \textbf{93.9} & \textbf{93.2} & 88.9 & 83.6 \\
  \end{tblr}}
  \caption{Effect of evaluating with text-conditioned image representations (\texttt{TCI}) on long-text retrieval instead of the standard \texttt{CLS} embedding.}
  \label{tab:tci}
\end{table}

%% file: tables/ablation_varying_cond_negatives.tex
\begin{table}[t]
  \centering\tiny
  \SetTblrInner{rowsep=0.25pt, colsep=1pt}
  \resizebox{\columnwidth}{!}{
  \begin{tblr}{
      colspec = {l !{\vrule width 0.1pt} c c c !{\vrule width 0.1pt} *{4}{c}!{\vrule width 0.1pt} *{4}{c} !{\vrule width 0.1pt} c},
      rows = {m},
      hline{1,Z} = {0.2pt},
      hline{3} = {0.2pt},
  }
  {Method} & & & & 
   \SetCell[c=4]{c} {FG-OVD} & & & &
   \SetCell[c=2]{c}{SV-1k} & &  \SetCell[c=2]{c}{U-1k} & &  \\
   \emph{ViT-B/16} & {C} & {S} & {P} & {Hard} & {Medium} & {Easy} & {Trivial} & 
  T2I & I2T &T2I & I2T & {Sim} \\
  \SetCell[c=13]{c}{\emph{$\beta$-CLIP (CE)}} \\
  \hline[0.2pt]
  \text{\tiny K=36, $\beta$=1} & - & - & - & \textbf{30.8} & \textbf{54.4 }& \textbf{59.0} & 80.6 & 93.4 & \textbf{94.1} & \textbf{88.7} & \textbf{88.7} & 0.98 \\
  \hline[0.2pt]
  \SetCell[c=13]{c}{\emph{+ Text-Conditioned Negatives}} \\
    \hline[0.2pt]
  K=36, & \checkmark & \checkmark &  & 28.3 & 48.1 & 52.9 & 79.9 & 93.7 & 94.0 & 88.4 & 88.5 & 0.98 \\
  K=36, &  & \checkmark & \checkmark & 28.9 & 52.3 & 55.4 & \textbf{80.9} & 93.6 & 93.9 & 87.6 & 88.3 & 0.97 \\
  K=36, & \checkmark & \checkmark & \checkmark & 27.3 & 50.1 & 54.0 & 80.6 & \textbf{93.8} & 93.6 & 87.7 & 87.9 & 0.97 \\
  \hline[0.2pt]
  \SetCell[c=13]{c}{\emph{$\beta$-CLIP (BCE)}} \\  
  \hline[0.2pt]
  \text{\tiny K=36, $\beta$=1} & - & - & - & 20.2 & 40.4 & 36.3 & 70.4 & \textbf{94.3} & 93.9 & 92.0 & 91.9 & 0.98 \\
  \hline[0.2pt]
  \SetCell[c=13]{c}{\emph{+ Text-Conditioned Negatives}} \\
    \hline[0.2pt]
  K=36, & \checkmark & \checkmark &  & 22.2 & 44.1 & 40.2 & \textbf{73.1} &\textbf{94.3} & 94.0 & 91.1 & 92.1 & 0.98 \\
  K=36, &  & \checkmark & \checkmark & 22.3 & 43.8 & 39.0 & 69.6 & \textbf{94.3} & 93.9 & \textbf{92.4} & 92.0 & 0.86 \\
  K=36, & \checkmark & \checkmark & \checkmark & \textbf{22.5} & \textbf{44.2} & \textbf{43.3} & 72.9 & 94.1 & \textbf{94.1} & 91.8 &\textbf{ 92.4} & 0.85 \\
  \end{tblr}}
  \vspace{1mm}
  \caption{Effectiveness of \textit{negative} hierarchical text conditioning. Columns “C”, “S”, and “P” correspond to Caption, Sentence, and Phrase inputs, respectively.
  }
  \label{tab:cond_negatives}
\end{table}

%% file: sec/6_conclusion.tex
\section{Conclusion}

In this work, we introduced $\beta$-CLIP, which aligns hierarchical text queries with text-conditioned visual features to enable fine-grained understanding without region supervision. To manage the resulting semantic overlaps, we proposed the $\beta$-Contextualized Alignment Loss ($\beta$-CAL), which balances precise self-matching with broader contextual consistency. Our experiments reveal that CE and BCE losses interact differently with hierarchical supervision: CE's softmax sharpens fine-grained discrimination, while BCE's sigmoid favors long-text retrieval. Overall, our findings demonstrate that explicitly modeling intra-image granularity provides a strong alternative to existing approaches for learning dense, fine-grained multimodal representations from long captions.

%% file: sec/X_suppl.tex
\clearpage
\setcounter{page}{1}

\maketitlesupplementary

\section{Experiments on ViT-L/14}
\label{sec:vit-l-14_exp}

We evaluate the scalability of $\beta$-CLIP using the larger CLIP ViT-L/14 @224px backbone, while keeping all other training settings fixed. Results are reported in Tables~\ref{tab:vit_l_fgovd} and~\ref{tab:vit_l_retrieval}.

\noindent\textbf{Fine-Grained Retrieval.} On FG-OVD, the CE variant benefits significantly from higher granularity ($K=36$), achieving 28.3–28.7\% on the Hard split (+13.0 over CLIP) and 70.9–72.2\% on Trivial (+32–33 over CLIP). We observe consistent gains of +8–10 when increasing $K$ from 6 to 36, validating the effectiveness of the soft-target CE formulation for multi-granular alignment. In this setting, $\beta=0.75$ consistently outperforms $\beta=0.5$ by a small margin. Interestingly, the BCE variant is not as effective on FG-OVD despite the larger backbone. 

Among methods that train on image–text pairs without region crops or hard negatives, $\beta$-CLIP CE achieves the highest performance, outperforming EVA-CLIP (18.3 Hard) and LongCLIP (9.6 Hard). It also performs better than FineCLIP (22.8 Hard), which uses cropped regions during training. The current state-of-the-art method, FG-CLIP (48.4 Hard), combines region crops with targeted hard-negative mining. Even so, $\beta$-CLIP CE, trained solely on decomposed long captions, reaches a significant fraction of its performance without explicit region level supervision.

\noindent\textbf{Long-Text Retrieval}
The pattern reverses on long-caption retrieval tasks between CE and BCE. Across the DCI, Urban1K, and ShareGPT4V-1K benchmarks, the BCE variant clearly outperforms CE, achieving 69.1–70.3\% T2I on the challenging DCI benchmark (versus 59.9–65.2\% for CE) and up to 94.9\% on Urban1K. This is consistent with our observations for the ViT-B/16 experiments which also show BCE's particular effectiveness for long-text image retrieval tasks. Increasing hierarchical granularity from K=6 to K=36 remains beneficial for CE on these tasks (+4–9 on DCI), as additional fine-grained phrases improve the fine-grained alignment of the nouns, actions, and spatial relations that are characteristic of DCI captions.

Across both ViT-B/16 and ViT-L/14 backbones, the results follow a consistent pattern. The CE variant outperforms BCE on fine-grained retrieval, whereas BCE excels at long-caption retrieval. In nearly all configurations, $\beta$=0.75 also yields small but reliable gains over $\beta$=0.5. This outcome reflects how $\beta$ moderates overlap among the intra-image positives. Increasing from 0.5 to 0.75 sufficiently increases useful contextual alignment from the multi-granular semantics.

\begin{table}[!h]
\centering
\caption{\textbf{Fine-Grained Retrieval Scalability (ViT-L/14).} Comparison on the FG-OVD benchmark. The Cross-Entropy (CE) variant significantly outperforms the Binary Cross-Entropy (BCE) variant. Performance consistently improves with higher hierarchical granularity ($K=36$) and stronger intra-image contextualization. ($\beta=0.75$). \textsuperscript{\textdaggerdbl} denotes FG-CLIP is trained with hard negatives.}
\label{tab:vit_l_fgovd}
\resizebox{0.75\linewidth}{!}{
\begin{tabular}{l|cccc}
\toprule
\multirow{2}{*}{{Model}} & \multicolumn{4}{c}{{FG-OVD}} \\
& {Hard} & {Medium} & {Easy} & {Trivial} \\
\midrule
CLIP & 15.4 & 25.3 & 25.7 & 38.8 \\
EVA-CLIP & 18.3 & 38.4 & 35.2 & 62.7 \\
LongCLIP & 9.6 & 19.7 & 16.0 & 39.8 \\
FineCLIP & 22.8 & 46.0 & 46.0 & \textbf{73.6} \\
\color{gray}FG-CLIP\textsuperscript{\textdaggerdbl}  & \color{gray}{48.4} & \color{gray}{69.5} &\color{gray}{ 71.2} & \color{gray}{89.7} \\
\midrule
\multicolumn{5}{c}{$\beta$-CLIP \textit{CE}} \\
\midrule
K=6, $\beta$=0.5 & 19.3 & 37.9 & 41.1 & 62.2 \\
K=36, $\beta$=0.5 & \underline{28.3} &  \underline{48.8} & \textbf{53.3} & 70.9 \\
\midrule
K=6, $\beta$=0.75 & 20.3 & 41.1 & 44.3 & 67.9 \\
K=36, $\beta$=0.75 & \textbf{28.7} & \textbf{49.1} &  \underline{52.1} &  \underline{72.2} \\

\midrule
\multicolumn{5}{c}{$\beta$-CLIP \textit{BCE}} \\
\midrule
K=6, $\beta$=0.5 & 15.1 & 34.6 & 29.2 & 32.4 \\
K=36, $\beta$=0.5 & 14.4 & 29.4 & 23.7 & 45.6 \\
\midrule 
K=6, $\beta$=0.75 & 15.0 & 36.3 & 33.1 & 36.8 \\
K=36, $\beta$=0.75 & 15.4 & 31.5 & 23.9 & 45.9 \\
\bottomrule
\end{tabular}
}
\end{table}

\begin{table}[!h]
\centering
\caption{\textbf{Long-Text Retrieval Scalability (ViT-L/14).} Comparison on the DCI, Urban1k, and ShareGPT4V-1K long-text retrieval benchmarks. In contrast to fine-grained tasks, the BCE variant consistently outperforms CE. \textsuperscript{\textdaggerdbl} denotes FG-CLIP is trained with hard negatives.}
\label{tab:vit_l_retrieval}
\resizebox{0.75\linewidth}{!}{
\begin{tabular}{l|cc|cc|cc}
\toprule
\multirow{2}{*}{{Model}} & \multicolumn{2}{c|}{{DCI}} & \multicolumn{2}{c|}{{Urban1k}} & \multicolumn{2}{c}{{ShareGPT4V}} \\
& {T2I} & {I2T} & {T2I} & {I2T} & {T2I} & {I2T} \\
\midrule
CLIP  & 36.4 & 37.2 & 90.7 & 89.2 & 83.6 & 86.5 \\
EVA-CLIP & 47.8 & 47.2 & - & - & 89.4 & 91.5 \\
LongCLIP & 52.5 & 44.2 & - & - & 95.6 & 95.8 \\
FineCLIP & 46.2 & 40.1 & - & - & 82.7 & 73.4 \\
TULIP & 56.4 & 55.7 & 91.1 & 90.1 & \textbf{99.0} & \textbf{99.0} \\
SmartCLIP & - & - & 90.1 & 93.0 & \underline{98.5} & \underline{97.9} \\
FineLIP & - & - & 93.9 & \underline{94.5} & - & - \\
\color{gray}FG-CLIP\textsuperscript{\textdaggerdbl}   & \color{gray}66.1 & \color{gray}66.7 & - & -  & \color{gray}96.8 & \color{gray}97.4\\
\midrule
\multicolumn{7}{c}{$\beta$-CLIP \textit{CE}} \\
\midrule
K=6, $\beta$=0.5 & 59.9 & 53.2 & 91.5 & 91.9 & 93.4 & 94.2 \\
K=36, $\beta$=0.5 & 65.2 & 62.4 & 93.2 & 93.2 & 94.1 & 94.3 \\
\midrule
K=6, $\beta$=0.75 & 60.3 & 55.7 & 91.9 & 92.3 & 93.4 & 94.2 \\
K=36, $\beta$=0.75 & 64.8 & 62.6 & 92.7 & 92.8 & 94.1 & 94.1 \\
\midrule
\multicolumn{7}{c}{$\beta$-CLIP \textit{BCE}} \\
\midrule
K=6, $\beta$=0.5 & \underline{69.7} & 67.0 & 94.3 & 93.7 & 94.1 & 94.3 \\
K=36, $\beta$=0.5 & 69.4 & \underline{68.2} & \underline{94.6} & \textbf{94.6} & 94.1 & 94.3 \\
\midrule
K=6, $\beta$=0.75 & 69.1 & 67.0 & 94.0 & 94.1 & 94.1 & 94.4 \\
K=36, $\beta$=0.75 & \textbf{70.3} & \textbf{68.5} & \textbf{94.9} & 94.3 & 94.2 & 94.3 \\
\bottomrule
\end{tabular}
}
\end{table}

\section{Larger Batch Sizes}
\label{sec:large_batch}

In contrastive vision-language pre-training, larger batch sizes are largely beneficial because they expose each image to more negative examples, which helps to improve global alignment. For $\beta$-CLIP, however, we note that a per-GPU batch size of $B$ does not correspond to a standard $B\times B$ similarity matrix. Because each image is paired with $K$ captions of varying granularity, each producing a text-conditioned image embedding, the resulting similarity matrix is of size $BK\times BK$. This already provides more contrastive signal than a standard CLIP setup with the same nominal batch size.

We ablate $\beta$-CLIP ViT-B/16 ($\beta$=0.5, K=6 or K=36) backbone on larger batch sizes from 64 to 96, and 112 (effective batch sizes 2048, 3072, 3584 respectively). As shown in Tables \ref{tab:bs_ce} and \ref{tab:bs_bce}, increasing batch size consistently degrades fine-grained retrieval performance, with the CE variant dropping up to 3.4 points on the Hard split, while long-caption retrieval remains relatively stable, occasionally showing minor gains or losses.

Additionally, the performance may also be effected due to asymmetric scaling of the two loss terms. The global CLS contrastive loss strengthens with batch size, benefiting disproportionately from the increased negatives for \textit{inter-image} discrimination. Although the fine-grained multi-granular loss also receives more cross-batch negatives, these additional negatives appear to be diluting the signal from the \textit{intra-image} positives. We speculate that this causes the optimization to prioritize separating distinct images over aligning the subtle, multi-granular intra-image features, similar to the effect of reducing $K$ in our method. Thus, given the already-expanded $BK\times BK$ similarity matrix, larger batches strengthen the coarse image-level objective at the expense of fine-grained correspondences.

\begin{table}[!h]
\centering
\caption{\textbf{Batch Size Ablation (CE Variants).} Effect of increasing batch size on retrieval performance using the ViT-B/16 backbone. Larger batch sizes result in a noticeable degradation in fine-grained retrieval (FG-OVD) capabilities, while global long-text retrieval remains relatively stable.}
\label{tab:bs_ce}
\resizebox{0.85\linewidth}{!}{
\begin{tabular}{c|cccc|cc|cc}
\toprule
\multirow{2}{*}{{BS}} & \multicolumn{4}{c|}{{FG-OVD}} &  \multicolumn{2}{c|}{{U-1K}} & \multicolumn{2}{c}{{SV-1K}} \\
& {Hard} & {Medium} & {Easy} & {Trivial} &  {T2I} & {I2T} & {T2I} & {I2T} \\
\midrule
\multicolumn{9}{c}{\textit{$\beta$-CLIP K=6, $\beta=0.5$}} \\
\midrule
64 & 29.2 & 51.3 & 56.4 & 81.5 & 87.9 & 88.4 & 93.5 & 94.0 \\
96 & 27.7 & 50.6 & 57.0 & 81.4 & 88.0 & 89.0 & 93.1 & 94.3 \\
112 & 25.8 & 48.9 & 54.8 & 80.3 & 88.4 & 89.5 & 93.7 & 94.3 \\
\midrule
\multicolumn{9}{c}{\textit{$\beta$-CLIP K=36, $\beta=0.5$}} \\
\midrule
64 & 30.9 & 55.4 & 60.4 & 80.3 & 89.0 & 88.6 & 93.7 & 94.0 \\
96 & 30.4 & 55.2 & 61.7 & 80.9 & 88.8 & 90.5 & 93.9 & 94.2 \\
112 & 28.3 & 53.3 & 58.8 & 80.4 & 89.4 & 89.2 & 94.1 & 94.1 \\
\bottomrule
\end{tabular}
}
\end{table}

\begin{table}[!h]
\centering
\caption{\textbf{Batch Size Ablation (BCE Variants).} Effect of increasing batch size for the BCE variant using the ViT-B/16 backbone. Increasing the batch size tends to negatively impact fine-grained performance (FG-OVD) while favoring global image-text alignment metrics.}
\label{tab:bs_bce}
\resizebox{0.75\linewidth}{!}{
\begin{tabular}{c|cccc|cc|cc}
\toprule
\multirow{2}{*}{{BS}} & \multicolumn{4}{c|}{{FG-OVD}} &  \multicolumn{2}{c|}{{U-1K}} & \multicolumn{2}{c}{{SV-1K}} \\
& {H} & {M} & {E} & {T} &  {T2I} & {I2T} & {T2I} & {I2T} \\
\midrule
\multicolumn{9}{c}{\textit{$\beta$-CLIP K=6, $\beta=0.5$}} \\
\midrule
64 & 20.6 & 42.7 & 45.3 & 71.2 & 91.8 & 92.0 & 94.5 & 94.0 \\
96  & 18.4 & 37.3 & 37.5 & 70.0 & 92.5 & 93.3 & 94.2 & 94.3 \\
112  & 17.0 & 33.9 & 31.9 & 71.1 & 92.6 & 92.8 & 94.2 & 94.3 \\
\midrule
\multicolumn{9}{c}{\textit{$\beta$-CLIP K=36, $\beta=0.5$}} \\
\midrule
64 & 20.1 & 38.5 & 34.2 & 71.3 & 91.8 & 92.3 & 94.4 & 94.1 \\
96 & 17.1 & 33.3 & 31.4 & 73.5 & 92.3 & 93.3 & 94.1 & 94.2 \\
112 & 14.8 & 27.6 & 24.7 & 71.0 & 92.7 & 93.0 & 94.0 & 94.1 \\
\bottomrule
\end{tabular}
}
\end{table}

\section{Distance-Calibrated Intra-Image Weights.}
Table~\ref{tab:ablation_distances} evaluates the effect of replacing uniform intra-image positive weights with exponentially distance-calibrated weights in $\beta$-CLIP. Across CE and BCE, and for both $K{=}6$ and $K{=}36$, the calibration yields only minor fluctuations in FG-OVD and long-text retrieval, with no consistent gains. These results suggest that once $\beta$ determines the overall contextualization strength, the precise allocation of mass across intra-image positives has limited effect, and uniform weighting is already sufficient.
\input{tables/ablation_varying_distances}

\section{Qualitative Analysis}
\label{sec:viz_analysis}

\begin{figure*}[t]
    \centering
    \renewcommand{\arraystretch}{0.0}
    \setlength{\tabcolsep}{1pt} 
    \begin{tabular}{@{}c ccc @{\hspace{2pt}}|@{\hspace{2pt}} ccc @{}}
    
    \rotatebox{90}{\parbox{2.2cm}{\centering Original Image}} &
    \includegraphics[width=0.138\textwidth]{figs/visualizations_final/originals/dog_nose_pos.png.png} &
    \includegraphics[width=0.138\textwidth]{figs/visualizations_final/originals/dog_nose_pos.png.png} &
    \includegraphics[width=0.138\textwidth]{figs/visualizations_final/originals/dog_nose_pos.png.png} &
    \includegraphics[width=0.138\textwidth]{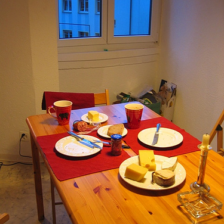} &
    \includegraphics[width=0.138\textwidth]{figs/visualizations_final_words/originals/2008_000184_cups_of_coffee_pos.png.png} &
    \includegraphics[width=0.138\textwidth]{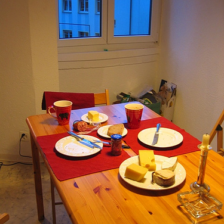}
    \\[1pt]
    
    \midrule[1pt] 
    \multicolumn{7}{c}{ \textbf{CLIP ViT-B/16 Baselines}} \\

    \\[1pt]
    
    \rotatebox{90}{\parbox{2.2cm}{\centering CLIP}} &
    \includegraphics[width=0.138\textwidth]{figs/visualizations_final/pretrained/dog_nose_pos.png} &
    \includegraphics[width=0.138\textwidth]{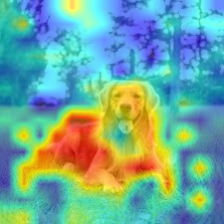} &
    \includegraphics[width=0.138\textwidth]{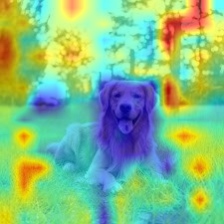} &
    \includegraphics[width=0.138\textwidth]{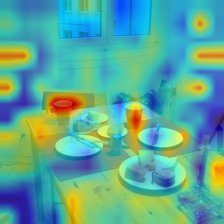} &
    \includegraphics[width=0.138\textwidth]{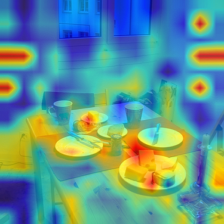} &
    \includegraphics[width=0.138\textwidth]{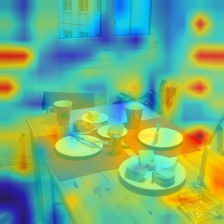}
    \\[-3pt]

    \rotatebox{90}{\parbox{2.2cm}{\centering CLIP \textbf{FT}}} &
    \includegraphics[width=0.138\textwidth]{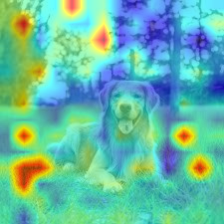} &
    \includegraphics[width=0.138\textwidth]{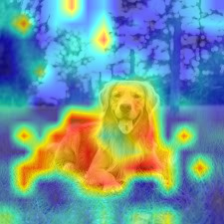} &
    \includegraphics[width=0.138\textwidth]{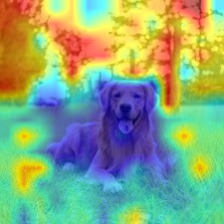} &
    \includegraphics[width=0.138\textwidth]{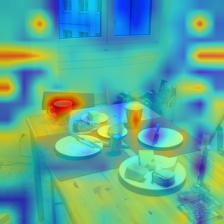} &
    \includegraphics[width=0.138\textwidth]{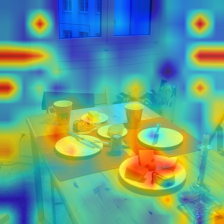} &
    \includegraphics[width=0.138\textwidth]{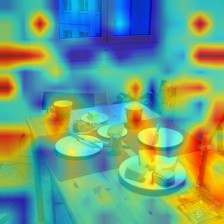}
    \\[1pt] 
    
    \midrule 
    \multicolumn{7}{c}{\textbf{$\beta$-CLIP ViT-B/16 Cross-Entropy (CE)}} \\
    \\[1pt]

    \rotatebox{90}{\parbox{2.2cm}{\centering \textbf{K=6}}} &
    \includegraphics[width=0.138\textwidth]{figs/visualizations_final/attn_pool+mlp_eos+5_cls+tcil_cs_k_positives_softmax_alpha_1_beta_0.5_bs64_ep10/dog_nose_pos.png} &
    \includegraphics[width=0.138\textwidth]{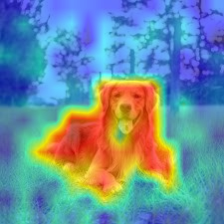} &
    \includegraphics[width=0.138\textwidth]{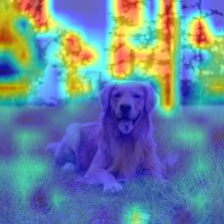} &
    \includegraphics[width=0.138\textwidth]{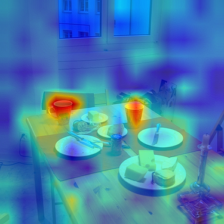} &
    \includegraphics[width=0.138\textwidth]{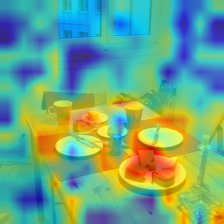} &
    \includegraphics[width=0.138\textwidth]{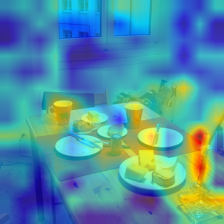}
    \\[-3pt]

    \rotatebox{90}{\parbox{2.2cm}{\centering \textbf{K=36}}} &
    \includegraphics[width=0.138\textwidth]{figs/visualizations_final/attn_pool+mlp_eos+5+concepts+30_cls+tcil_csk_k_positives_softmax_alpha_1_beta_0.5_bs64_ep10/dog_nose_pos.png} &
    \includegraphics[width=0.138\textwidth]{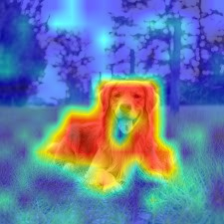} &
    \includegraphics[width=0.138\textwidth]{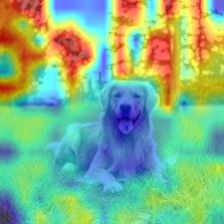} &
    \includegraphics[width=0.138\textwidth]{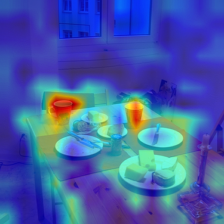} &
    \includegraphics[width=0.138\textwidth]{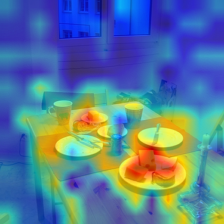} &
    \includegraphics[width=0.138\textwidth]{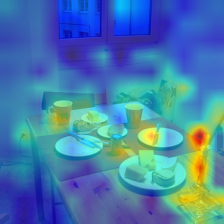}
    \\[1pt]

    \midrule 
    \multicolumn{7}{c}{ \textbf{$\beta$-CLIP ViT-B/16 Binary-Cross-Entropy (BCE)}} \\
    \\[1pt]

    \rotatebox{90}{\parbox{2.2cm}{\centering \textbf{K=6}}} &
    \includegraphics[width=0.138\textwidth]{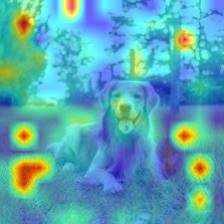} &
    \includegraphics[width=0.138\textwidth]{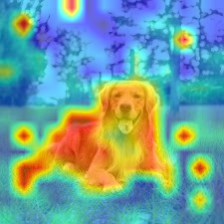} &
    \includegraphics[width=0.138\textwidth]{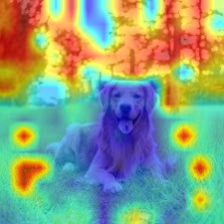} &
    \includegraphics[width=0.138\textwidth]{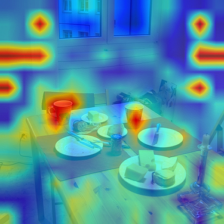} &
    \includegraphics[width=0.138\textwidth]{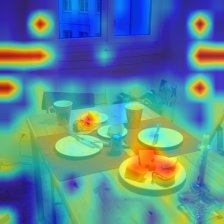} &
    \includegraphics[width=0.138\textwidth]{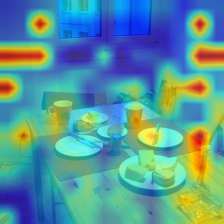}
    \\[1pt]
    
    \rotatebox{90}{\parbox{2.2cm}{\centering \textbf{K=36}}} &
    \includegraphics[width=0.138\textwidth]{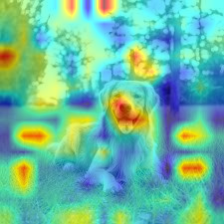} &
    \includegraphics[width=0.138\textwidth]{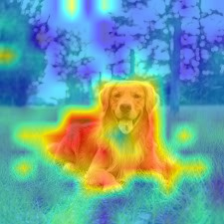} &
    \includegraphics[width=0.138\textwidth]{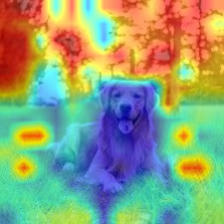} &
    \includegraphics[width=0.138\textwidth]{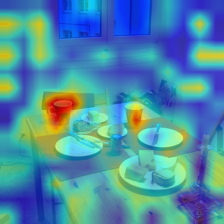} &
    \includegraphics[width=0.138\textwidth]{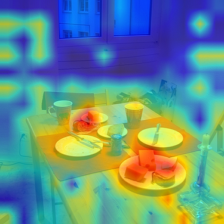} &
    \includegraphics[width=0.138\textwidth]{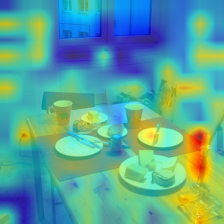}
    \\[-3pt]
    
    \\[6pt]
    &
    {\parbox{0.138\textwidth}{\centering \texttt{nose}}} &
    { \parbox{0.138\textwidth}{\centering \texttt{dog}}} &
    { \parbox{0.138\textwidth}{\centering \texttt{trees}}} &
    { \parbox{0.138\textwidth}{\centering \texttt{cups of coffee}}} &
    { \parbox{0.138\textwidth}{\centering \texttt{cheese}}} &
    { \parbox{0.138\textwidth}{\centering \texttt{candle}}}
    \\[-2pt]
    \end{tabular}
    \caption{\textbf{Heatmaps of Patch-Text Logit Similarities for Short Phrases.} Rows 1-2: Original CLIP and CLIP FT exhibit diffuse similarity patterns, characterized by frequent secondary peaks on semantically unrelated regions such as the grass or background walls. Rows 3-6: $\beta$-CLIP successfully disentangles these fine-grained concepts. CE yields the most spatially sparse distributions, concentrating similarity primarily on the regions most relevant to the text. BCE shows higher similarity with background features, comparable to the baselines at lower granularity, K=6. However, as $K$ increases, these irrelevant regions are more effectively suppressed, and the similarities become progressively more concentrated on the semantically relevant regions.}
    \label{fig:viz_dog_table}
\end{figure*}

\begin{figure*}[t]
    \centering
    \renewcommand{\arraystretch}{0.0}
    \setlength{\tabcolsep}{1pt} 
    \begin{tabular}{@{}c cc @{\hspace{2pt}}|@{\hspace{2pt}} cc @{\hspace{2pt}}|@{\hspace{2pt}} cc @{}}
    
    \rotatebox{90}{\parbox{2.2cm}{\centering  Original Image}} &
    \includegraphics[width=0.122\textwidth]{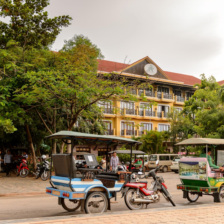} &
    \includegraphics[width=0.122\textwidth]{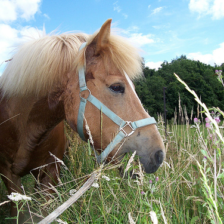} &
    \includegraphics[width=0.122\textwidth]{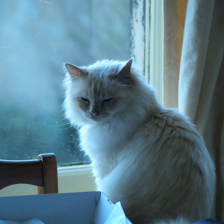} &
    \includegraphics[width=0.122\textwidth]{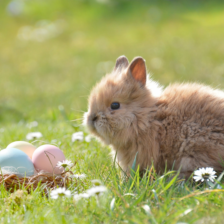} &
    \includegraphics[width=0.122\textwidth]{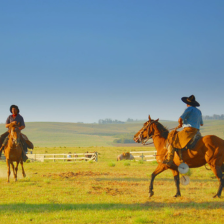} &
    \includegraphics[width=0.122\textwidth]{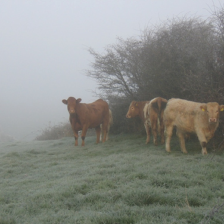}
    \\[1pt]
    
    \midrule[1pt] 
    \multicolumn{7}{c}{ \textbf{CLIP ViT-B/16 Baselines}} \\
    \\[1pt]
    
    \rotatebox{90}{\parbox{2.2cm}{\centering CLIP}} &
    \includegraphics[width=0.122\textwidth]{figs/visualizations_final/pretrained/sa_1544018_Despite_the_background_the_colorful_tuk-tuks_and_chatting_locals_stand_out_in_this_image._pos.png} &
    \includegraphics[width=0.122\textwidth]{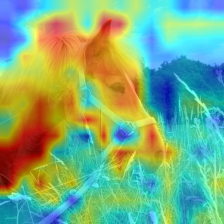} &
    \includegraphics[width=0.122\textwidth]{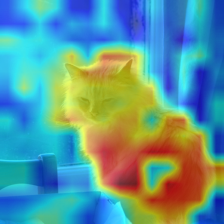} &
    \includegraphics[width=0.122\textwidth]{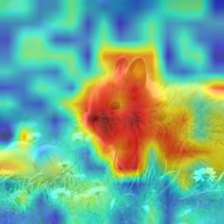} &
    \includegraphics[width=0.122\textwidth]{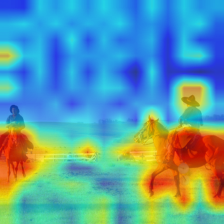} &
    \includegraphics[width=0.122\textwidth]{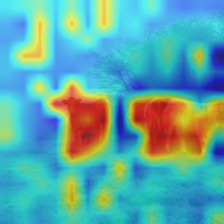}
    \\[-3pt]

    \rotatebox{90}{\parbox{2.2cm}{\centering CLIP \textbf{FT}}} &
    \includegraphics[width=0.122\textwidth]{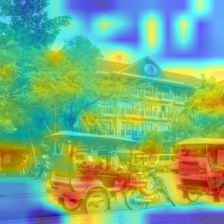} &
    \includegraphics[width=0.122\textwidth]{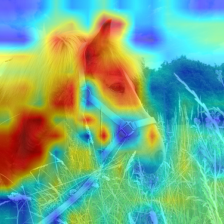} &
    \includegraphics[width=0.122\textwidth]{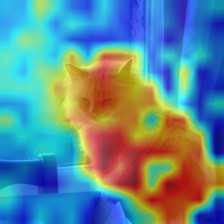} &
    \includegraphics[width=0.122\textwidth]{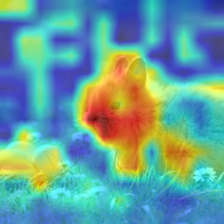} &
    \includegraphics[width=0.122\textwidth]{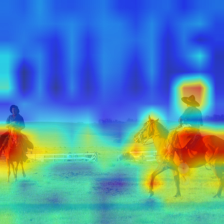} &
    \includegraphics[width=0.122\textwidth]{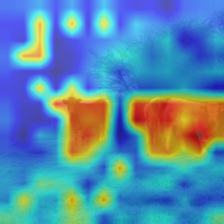}
    \\[1pt] 
    
    \midrule
    \multicolumn{7}{c}{\textbf{$\beta$-CLIP ViT-B/16 Cross-Entropy (CE)}} \\
    \\[1pt]

    \rotatebox{90}{\parbox{2.2cm}{\centering \textbf{K=6}}} &
    \includegraphics[width=0.122\textwidth]{figs/visualizations_final/attn_pool+mlp_eos+5_cls+tcil_cs_k_positives_softmax_alpha_1_beta_0.5_bs64_ep10/sa_1544018_Despite_the_background_the_colorful_tuk-tuks_and_chatting_locals_stand_out_in_this_image._pos.png} &
    \includegraphics[width=0.122\textwidth]{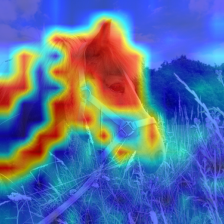} &
    \includegraphics[width=0.122\textwidth]{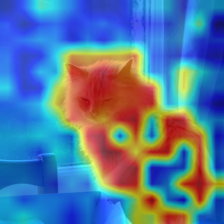} &
    \includegraphics[width=0.122\textwidth]{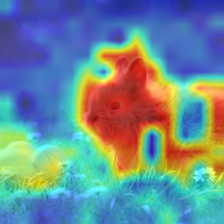} &
    \includegraphics[width=0.122\textwidth]{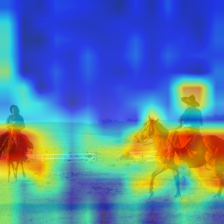} &
    \includegraphics[width=0.122\textwidth]{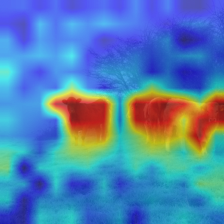}
    \\[-3pt]

    \rotatebox{90}{\parbox{2.2cm}{\centering \textbf{K=36}}} &
    \includegraphics[width=0.122\textwidth]{figs/visualizations_final/attn_pool+mlp_eos+5+concepts+30_cls+tcil_csk_k_positives_softmax_alpha_1_beta_0.5_bs64_ep10/sa_1544018_Despite_the_background_the_colorful_tuk-tuks_and_chatting_locals_stand_out_in_this_image._pos.png} &
    \includegraphics[width=0.122\textwidth]{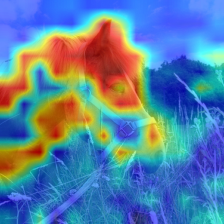} &
    \includegraphics[width=0.122\textwidth]{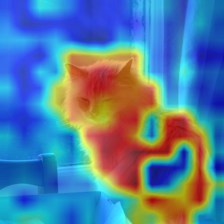} &
    \includegraphics[width=0.122\textwidth]{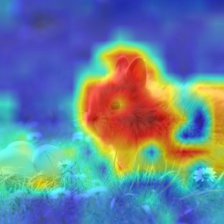} &
    \includegraphics[width=0.122\textwidth]{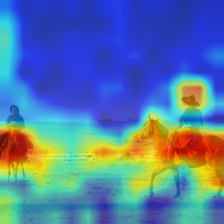} &
    \includegraphics[width=0.122\textwidth]{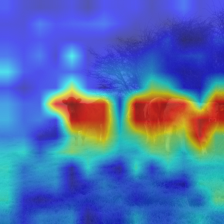}
    \\[1pt]

    \midrule
    \multicolumn{7}{c}{ \textbf{$\beta$-CLIP ViT-B/16 Binary-Cross-Entropy (BCE)}} \\
    \\[1pt]

    \rotatebox{90}{\parbox{2.2cm}{\centering \textbf{K=6}}} &
    \includegraphics[width=0.122\textwidth]{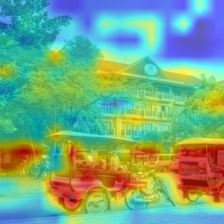} &
    \includegraphics[width=0.122\textwidth]{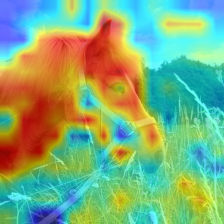} &
    \includegraphics[width=0.122\textwidth]{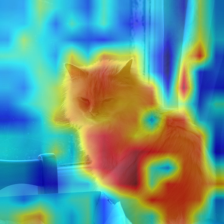} &
    \includegraphics[width=0.122\textwidth]{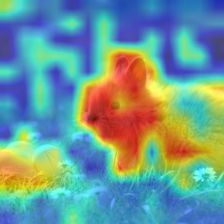} &
    \includegraphics[width=0.122\textwidth]{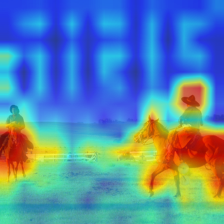} &
    \includegraphics[width=0.122\textwidth]{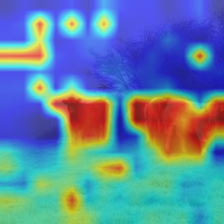}
    \\[-3pt]
    
    \rotatebox{90}{\parbox{2.2cm}{\centering \textbf{K=36}}} &
    \includegraphics[width=0.122\textwidth]{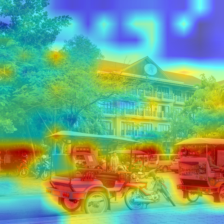} &
    \includegraphics[width=0.122\textwidth]{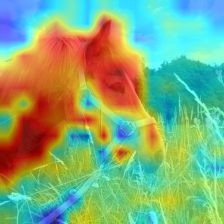} &
    \includegraphics[width=0.122\textwidth]{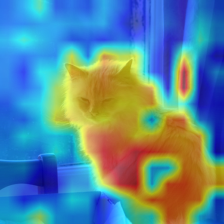} &
    \includegraphics[width=0.122\textwidth]{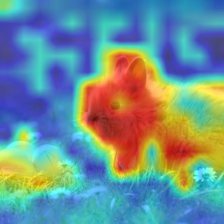} &
    \includegraphics[width=0.122\textwidth]{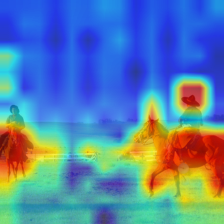} &
    \includegraphics[width=0.122\textwidth]{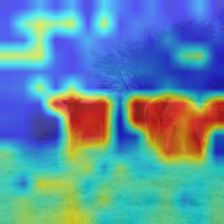}
    \\[-3pt]
    
    \\[6pt]
    &
    {\parbox{0.122\textwidth}{\centering Despite the background the \underline{colorful tuk-tuks} and \underline{chatting locals} stand out in this image.}} &
    { \parbox{0.122\textwidth}{\centering The \underline{pointed ears} of a \underline{horse}, alert and listening to the surroundings.}} &
    { \parbox{0.122\textwidth}{\centering A fluffy \underline{white cat} sitting upright.}} &
    { \parbox{0.122\textwidth}{\centering A fluffy and adorable \underline{bunny rabbit}.}} &
    { \parbox{0.122\textwidth}{\centering \underline{Several horses} being ridden by \underline{cowboys} in an outdoor \underline{western setting}}} &
    { \parbox{0.122\textwidth}{\centering Friendly \underline{cows} hanging out and being super chill.}}
    \\[-2pt]
    \end{tabular}
    \caption{\textbf{Heatmaps of Patch-Text Logit Similarities for Long Captions.} Rows 1-2: Original CLIP and CLIP FT exhibit diffuse similarity patterns, assigning high similarity to broader background regions. Rows 3-6: $\beta$-CLIP improves alignment with complex semantics. CE yields the sharpest localization, effectively isolating specific details (ex., the ``pointed ears'') and increasingly so with higher K (ex. ``western setting'' at K=36). BCE captures a wider semantic scope, grounding multi-instance concepts like the ``chatting locals,'' better than the rest. Increasing $K$ suppresses similarity to irrelevant regions (most evident in the cowboys and cows scenes). Notably, all models exhibit sensitivity to salient unmentioned objects that contextually co-occur with the query, such as the eggs next to the bunny.}
    \label{fig:viz_all}
\end{figure*}

\begin{figure*}[t]
    \centering
    \renewcommand{\arraystretch}{0.0}
    \setlength{\tabcolsep}{2pt} 
    
    \begin{tabular}{@{}c cc @{}}    
    \rotatebox{90}{\parbox{1.4cm}{\centering  Original Image}} &
    \multicolumn{2}{c}{
        \includegraphics[width=0.125\textwidth]{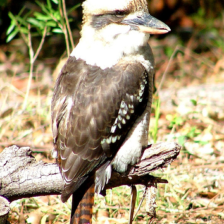}
    }
    \\[1pt]
    
    \midrule[0.5pt] 
    \multicolumn{3}{c}{ \textbf{CLIP ViT-B/16 Baselines}} \\
    \midrule[0.5pt] 
    
    \rotatebox{90}{\parbox{1.4cm}{\centering  CLIP}} &
    \includegraphics[width=0.125\textwidth]{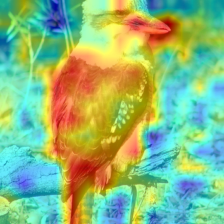} &
    \includegraphics[width=0.125\textwidth]{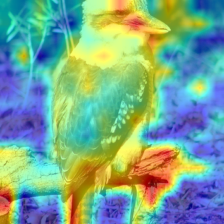} 
    \\[0pt]

    \rotatebox{90}{\parbox{1.4cm}{\centering  CLIP \textbf{FT}}} &
    \includegraphics[width=0.125\textwidth]{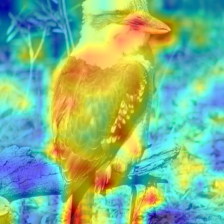} &
    \includegraphics[width=0.125\textwidth]{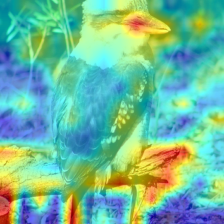}
    \\[1pt] 
    
    \midrule[0.5pt]
    \multicolumn{3}{c}{\textbf{$\beta$-CLIP ViT-B/16 CE}} \\
    \midrule[0.5pt] 

    \rotatebox{90}{\parbox{1.4cm}{\centering  \textbf{K=6}}} &
    \includegraphics[width=0.125\textwidth]{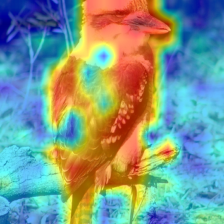} &
    \includegraphics[width=0.125\textwidth]{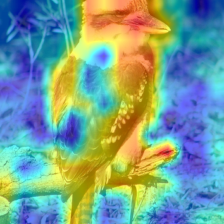} 
    \\[0pt]

    \rotatebox{90}{\parbox{1.4cm}{\centering  \textbf{K=36}}} &
    \includegraphics[width=0.125\textwidth]{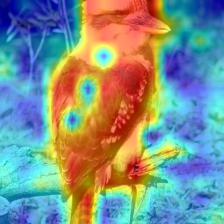} &
    \includegraphics[width=0.125\textwidth]{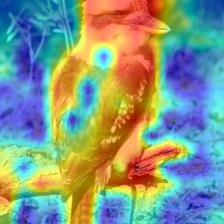} 
    \\[1pt]

    \midrule[0.5pt]
    \multicolumn{3}{c}{ \textbf{$\beta$-CLIP ViT-B/16 BCE}} \\
    \midrule[0.5pt] 

    \rotatebox{90}{\parbox{1.4cm}{\centering  \textbf{K=6}}} &
    \includegraphics[width=0.125\textwidth]{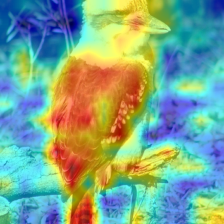} &
    \includegraphics[width=0.125\textwidth]{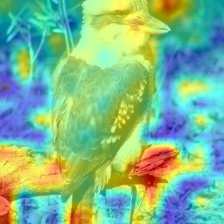} 
    \\[0pt]
    
    \rotatebox{90}{\parbox{1.4cm}{\centering  \textbf{K=36}}} &
    \includegraphics[width=0.125\textwidth]{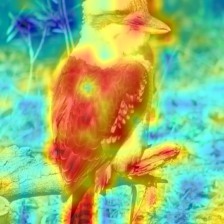} &
    \includegraphics[width=0.125\textwidth]{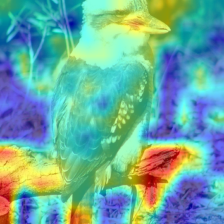}
    \\[-2pt]
    
    \\[2pt]
    &
    {\parbox{0.125\textwidth}{\centering The brown and white \underline{feathered body} of a \underline{bird}.}} &
    {\parbox{0.125\textwidth}{\centering A stout \underline{bird} sitting on a \underline{textured log}.}} 
    \\[-2pt]
    \end{tabular}
    \caption{\textbf{Heatmaps of Patch-Text Logit Similarities: CE vs BCE.} Rows 1-2: Original CLIP and CLIP FT exhibit diffuse similarity patterns with a disproportionate focus on the beak, regardless of the query. Rows 3-6: $\beta$-CLIP comparisons reveal distinct localization behaviors. CE provides significantly higher coverage of the bird's body, and increasingly with higher $K$. BCE is unable to localize the entire body given the word ``bird'' in a different context; instead, it focuses strictly on salient features like the beak and tail. However, given a more detailed description, such as ``feathered body,'' it is able to achieve better localization. Both methods effectively ground the distinct background element (``textured log''), with boundary precision improving at higher $K$.}
    \label{fig:viz_all_2}
\end{figure*}

We visualize patch-text logit similarities across distinct scenarios: short phrases isolation (Fig. \ref{fig:viz_dog_table}) and long-caption grounding (Fig. \ref{fig:viz_all} and Fig. \ref{fig:viz_all_2}). Across these settings, distinct trends emerge when comparing the baselines to the Cross-Entropy (CE) and Binary Cross-Entropy (BCE) variants of $\beta$-CLIP.

\noindent\textbf{Baseline Saliency}
Both the pre-trained CLIP and CLIP fine-tuned on long-text (CLIP FT) exhibit a strong \textit{saliency bias}. Rather than localizing the specific text query, they tend to focus disproportionately on the most visually salient or high-contrast features. This is most evident in the Fig. \ref{fig:viz_all_2}, where baselines consistently focus on the bird's beak even when queried about ``wings" or ``body." Similarly, in short-concept scenarios (Fig. \ref{fig:viz_dog_table}), baseline similarities frequently encompass unrelated background regions like grass, sky, or walls, showing a reliance on global image features rather than explicit region-text correspondences.

\noindent\textbf{Semantic Disentanglement}
A primary distinction between our proposed loss variants is the density of their heatmaps and their ability to separate semantically related concepts.
\begin{itemize}
    \item \textbf{Cross-Entropy (CE):} The CE variant yields highly sparse and sharp localizations. In Fig. \ref{fig:viz_dog_table}, CE effectively suppresses similarity to background regions, localizing objects like ``nose", ``candle", and the fence for ``western setting" with high precision.
    \item \textbf{Binary Cross-Entropy (BCE):} The BCE variant maintains a broader semantic scope. While this results in higher initial background noise (e.g., the curtain in Fig. \ref{fig:viz_all}), it allows the model to capture multi-instance concepts better (e.g., ``chatting locals" in Fig. \ref{fig:viz_all}).
\end{itemize}

\noindent\textbf{The Effect of Granularity ($K$).}
For both variants, but particularly for BCE, increasing the size of the hierarchy $K$ suppresses noisy features better. At lower K, BCE exhibits diffuse similarities comparable to the baselines. However, as $K$ increases to 36, these diffuse regions become noticeably less. This effectively suppresses irrelevant background regions (ex. the ``cowboys" and ``cows" in Fig. \ref{fig:viz_all}) and sharpens object boundaries (ex. textured log in Fig. \ref{fig:viz_all_2}. It is additionally able to localize fine-grained semantics such as the fence with increasing precision as K incrases (ex. ``western setting" in Fig. \ref{fig:viz_all}).

\section{Implementation Details}
\label{sec:implementation}
Our approach introduces a randomly initialized Cross-Attention Transformer Block (8 heads, 512 hidden dim, MLP expansion ratio 4, Pre-Norm) to the standard CLIP model. The CLIP vision and text encoders are initialized from OpenAI's pre-trained weights. We fine-tune the model on the filtered ShareGPT4V-1.2M dataset. Table~\ref{tab:datasets_stats} summarizes key statistics of the data used during training and evaluation. Training is done using 4 NVIDIA A100 (80GB) GPUs. The optimizer settings, learning‑rate schedule and data‑augmentation pipeline used during fine‑tuning are described in Table \ref{tab:supp_hparams}.

\begin{table}[h]
    \centering
    \footnotesize
    \setlength{\tabcolsep}{4pt}
    \caption{Hyper-parameters for $\beta$-CLIP}
    \label{tab:supp_hparams}
    \begin{tabular}{l c}
        \toprule
        \textbf{Parameter} & \textbf{Value} \\
        \midrule
        Epochs                  & 10 \\
        Batch size              & 2,048 \\
        Optimizer               & AdamW $(\beta_1{=}0.9,\ \beta_2{=}0.98)$ \\
        Initial LR              & $1{\times}10^{-9}$ \\
        LR                      & $1{\times}10^{-5}$ \\
        LR Cross-Attn           & $1{\times}10^{-3}$ \\
        Final LR Cross-Attn     & $1{\times}10^{-4}$ \\
        LR schedule             & cosine decay \\
        Weight decay            & 0.01 \\
        Warm-up epochs          & 0.1 \\
        \addlinespace[0.3em]
        Image Augmentation     & \begin{tabular}[c]{@{}l@{}}
                                  RandomResizedCrop (0.5–1.0) \\
                                  Normalize: \\
                                  $\mu{=}(0.481, 0.458, 0.408)$ \\
                                  $\sigma{=}(0.269, 0.261, 0.276)$
                                \end{tabular} \\
        \bottomrule
    \end{tabular}
\end{table}

\subsection{Multi-Granular Data Setup}

\textbf{Caption Filtering:} We use Detoxify \cite{detoxify} and FalconsAI \cite{falconsai} to filter out content with a toxicity score $>0.1$ and $>0.5$ respectively. These samples include approximately $2K$ image-text pairs containing zero-tolerance content such as child abuse and pornography.
\label{supp:filtering}

\noindent\textbf{Caption Preprocessing:} We use regex pattern matching to identify and remove repeated substrings (e.g., character-level repetitions) and `itertools.groupby` to eliminate consecutive word-level repetitions, which are common in generated long captions.

\noindent\textbf{Hierarchical Parsing:} To generate the multi-granular queries, we utilize the \texttt{spaCy} library with the en\_core\_web\_sm model.

\begin{enumerate}
    \item \textbf{Sentences:} The caption is split using standard sentence tokenization. During training, we sample $N$ sentences (without replacement if enough sentences are available) to serve as coarse-grained queries.
    \item \textbf{Phrases:} We extract fine-grained phrases using a custom extractor:
    \begin{itemize}
        \item \textbf{Noun Chunks:} We extract base noun phrases and extend them to include spatial indicators (e.g., ``...on the left") using custom matchers.
        \item \textbf{Actions:} We identify action-oriented phrases by matching \texttt{VERB+ADP} patterns (e.g., ``leaning against").
        \item \textbf{Spatial Relations:} We parse spatial prepositional phrases anchored by a set of predefined \texttt{SPATIAL HEAD} tokens (including directions like \textit{left, right, top, bottom} and positions like \textit{center, middle, near}). We extract phrases matching the pattern \texttt{ADP [DET]? SPATIAL HEAD [of]?}, capturing relative positions such as to the left of'' or in the center,'' alongside explicit adverbial relations like ``next to.''
    \end{itemize}
\end{enumerate}
We filter out phrases shorter than 3 characters or those consisting solely of stop words to ensure semantic meaningfulness.

\vspace{0.5cm}

\begin{table}[h]
\centering
\footnotesize
\setlength{\tabcolsep}{3.5pt}
\caption{Statistics of the training corpus and evaluation benchmarks.}
\label{tab:datasets_stats}
\begin{tabular}{l c c c}
      \toprule
Dataset & Task &Images & Text / Query \\ \midrule
ShareGPT4V-1.2M \cite{chen2024sharegpt4v}      & Training & 1.2M            & - \\

Flickr30k \cite{young2014imageflickr30k}           & Coarse-Grained          & 31k          & 5 cap $\times$ img \\
MS-COCO14 (5k) \cite{lin2014microsoftmscoco}     & Coarse-Grained           & 5k           & 5 cap $\times$ img \\ 
DCI \cite{urbanek2024dci}                & Long-Text & 8k            & 1 cap $\times$ img \\
Urban1k  \cite{zhang2024longclip}           & Long-Text & 1k            & 1 cap $\times$ img \\
ShareGPT4V-1K \cite{chen2024sharegpt4v}      & Long-Text & 1k            & 1 cap $\times$ img \\
FG-OVD (ex. Hard) \cite{bianchi2023devil-cdf}  & Fine-grained & 1.7k & 2.3k positive cap \\ 
                    &                        &      & $+$\,23k neg cap \\ 
\bottomrule
\end{tabular}
\end{table}

%% file: tables/ablation_varying_distances.tex
\begin{table}[!h]
  \centering\tiny
  \SetTblrInner{rowsep=0.25pt, colsep=1pt}
  \resizebox{0.85\columnwidth}{!}{
  \begin{tblr}{
      colspec = {l *{4}{c} !{\vrule width 0.1pt} *{4}{c} !{\vrule width 0.1pt} c},
      rows = {m},
      row{1-2} = {},
      column{1} = {halign=l},
      column{2-Z} = {halign=c},
      hline{1,Z} = {0.2pt},
      hline{3} = {0.2pt},
      hline{2}={6-7}{leftpos=-1,rightpos=-1,endpos},
      hline{2}={8-9}{leftpos=-1,rightpos=-1,endpos},
      cell{1}{6} = {c=2}{c},
      cell{1}{8} = {c=2}{c},
  }
    {Method} & \SetCell[c=4]{c} {FG-OVD} & & & & {SV-1k} & & {U-1k} & &  \\
   \emph{ViT-B/16} & {Hard} & {Medium} & {Easy} & {Trivial} & {T2I} & {I2T} & {T2I} & {I2T} & {Sim}  \\
  \SetCell[c=10]{c}{\emph{$\beta$-CLIP (CE)}} \\
  \hline[0.2pt]
  \text{\tiny K=6, $\beta$=0.75} & 30.6 & 52.6 & 58.5 & 82.1 & 93.4 & 93.9 & 87.3 & 88.7 & 0.97  \\
  \text{\tiny K=36, $\beta$=0.75} & 30.7 & 54.2 & 60.0 & 80.5 & 93.7 & 94.1 & 88.6 & 88.7 & 0.98 \\
  \hline[0.2pt]
  \SetCell[c=10]{c}{\emph{+ distance-calibrated weights}} \\
  \hline[0.2pt]

  \text{\tiny K=6, $\beta$=0.75} & 29.9 & 51.7 & 58.2 & 81.4 & 93.3 & 94.1 
  & 88.1 & 88.2 & 0.97 \\
  \text{\tiny K=36, $\beta$=0.75} & 30.3 & 54.3 & 60.1 & 80.6 & 93.6 & 94.1 & 89.1 & 88.3 & 0.98 \\
  \hline[0.2pt]
  \SetCell[c=10]{c}{\emph{$\beta$-CLIP (BCE)}} \\
  \hline[0.2pt]
  \text{\tiny K=6, $\beta$=0.75} & 20.6 & 42.4 & 45.7 & 71.8 & 94.3 & 94.0 & 91.7 & 91.4 & 0.94 \\
  \text{\tiny K=36, $\beta$=0.75} & 19.8 & 38.0 & 34.2 & 72.8 & 94.3 & 93.8 & 91.8 & 91.8 & 0.98 \\
  \hline[0.2pt]
  \SetCell[c=10]{c}{\emph{+ distance-calibrated weights}} \\
  \hline[0.2pt]
  \text{\tiny K=6, $\beta$=0.75} & 20.1 & 41.8 & 45.6 & 71.3 & 94.4 & 94.4 & 91.3 & 91.5 & 0.95 \\
  \text{\tiny K=36, $\beta$=0.75} & 21.2 & 40.5 & 36.3 & 72.6 & 94.3 & 94.1 & 91.6 & 91.4 & 0.98 \\
  \end{tblr}}
  \caption{Effect of distance-calibrated intra-image positive weights at $\beta{=}0.75$. Calibrated targets decay with scale distance (e.g., for $K{=}6$: $1.00, 0.71, 0.68, 0.65 \ldots$), in contrast to the uniform BCE weights ($1.00, 0.75, 0.75 0.75 \ldots$)}
  \label{tab:ablation_distances}
\end{table}